\documentclass{article}

\newcommand{\ind}{\mathrm{Ind}}
\newcommand{\ksma}{\textsc{$k$SM-Approx}}

\usepackage[letterpaper,top=2cm,bottom=2cm,left=3cm,right=3cm,marginparwidth=1.75cm]{geometry}
\usepackage{wrapfig}

\usepackage[toc,page]{appendix}
\usepackage{amsmath}
\usepackage[colorlinks=true, allcolors=blue]{hyperref}
\usepackage{bm}

\usepackage{fullpage}
\usepackage{amsthm,amsfonts}
\usepackage[bb=boondox]{mathalfa}

\usepackage[utf8]{inputenc}
\usepackage[english]{babel}
\usepackage{amsmath,amssymb,amsfonts}
\makeatletter
\let\NAT@parse\undefined
\makeatother
\PassOptionsToPackage{hyphens}{url}\usepackage[colorlinks=true, allcolors=blue]{hyperref}
\usepackage[linesnumbered,lined,ruled,vlined,noend]{algorithm2e}
\usepackage{algpseudocode}
\usepackage{xfrac}
\usepackage[caption=false,font=footnotesize]{subfig}
\usepackage[bb=boondox]{mathalfa}
\usepackage{nameref}
\usepackage{mathtools}
\usepackage{float}
\newcommand{\tl}{z}

\DeclareMathAlphabet{\mathpzc}{OT1}{pzc}{m}{it}
\newcommand{\sub}{S}
\newcommand{\pathalg}{\textsc{Central-Path}}
\newcommand{\iter}{t}
\newcommand{\dd}{\mathcal{D}}
\newcommand{\zz}{z}

\newcommand{\abs}[1]{\left| #1\right|}
\renewcommand{\paragraph}[1]{\medskip\noindent\textbf{{#1}}}
\newcommand{\dist}[1]{\mathrm{dist}\rbr{#1}}
\newcommand{\diag}[1]{\mathrm{diag}\rbr{#1}}
\newtheorem{theorem}{Theorem}[section]
\newtheorem{lemma}[theorem]{Lemma}
\newtheorem{corollary}{Corollary}[theorem]
\newtheorem{proposition}[theorem]{Proposition}

\newcommand{\ksm}{\mathrm{ksm}}

\newtheorem{definition}[theorem]{Definition}

\newcommand{\rlksm}{\mathrm{relax\ksm}}
\newcommand{\REAL}{\mathbb{R}}

\newcommand{\psd}{\mathbb{S}^+}
\newcommand{\soc}{\mathbb{Q}^+}
\newcommand{\ipsd}{\mathbb{S}^{++}}
\newcommand{\isoc}{\mathbb{Q}^{++}}
\newcommand{\doubleOne}{\mathbb{1}}
\newcommand{\KSM}{{\text{$k$SM}}}
\newcommand{\norm}[1]{\left\lVert#1\right\rVert}

\newcommand{\br}[1]{\left\{#1\right\}}
\newcommand{\rbr}[1]{\left(#1\right)}
\newcommand{\cbr}[1]{\left[#1\right]}
\newcommand{\eps}{\ensuremath{\varepsilon}}

\newcommand{\poly}[0]{\mathrm{poly}}
\newcommand{\nnz}[1]{\mathrm{nnz}\rbr{#1}}
\newcommand{\rank}[1]{\mathrm{rank}\rbr{#1}}
\newcommand{\trace}[1]{\mathrm{trace}\rbr{#1}}

\newcommand{\interior}[1]{\mathbf{int}\rbr{#1}}
\newcommand*{\rom}[1]{\uppercase\expandafter{\romannumeral #1\relax}}
\newcommand{\comment}[1]{}
\let\oldnl\nl
\newcommand{\nonl}{\renewcommand{\nl}{\let\nl\oldnl}}

\title{$k$-PCA for (non-squared) Euclidean Distances:\\ Polynomial Time Approximation}
\author {
    Daniel Greenhut\textsuperscript{$\dagger$}, Dan Feldman\\
    Robotics \& Big Data Labs, University of Haifa, Israel\\
    {\tt\small \{daniel.greenhut, dannyf.post\}@gmail.com}
    \thanks{\textsuperscript{$\dagger$}This paper is in partial fulfillment of the requirements for the degree of master of science in the Computer
    Science Department, University of Haifa.}
}

\date{\today}
\begin{document}

\maketitle
\thispagestyle{empty}
\pagestyle{empty}

\begin{abstract}
Given an integer $k\geq1$ and a set $P$ of $n$ points in $\REAL^d$, the classic $k$-PCA (Principle Component Analysis) approximates the affine \emph{$k$-subspace mean} of $P$, which is the $k$-dimensional affine linear subspace that minimizes its sum of squared Euclidean distances ($\ell_{2,2}$-norm) over the points of $P$, i.e., the mean of these distances.
The \emph{$k$-subspace median} is the subspace that minimizes its sum of (non-squared) Euclidean distances ($\ell_{2,1}$-mixed norm), i.e., their median. The median subspace is usually more sparse and robust to noise/outliers than the mean, but also much harder to approximate since, unlike the $\ell_{z,z}$ (non-mixed) norms, it is non-convex for $k<d-1$.

We provide the first polynomial-time deterministic algorithm whose both running time and approximation factor are not exponential in $k$. More precisely, the multiplicative approximation factor is $\sqrt{d}$, and the running time is polynomial in the size of the input. We expect that our technique would be useful for many other related problems, such as $\ell_{2,z}$ norm of distances for $z\not \in \br{1,2}$, e.g., $z=\infty$, and handling outliers/sparsity.

Open code and experimental results on real-world datasets are also provided.
\end{abstract}

\section{Introduction}
\label{background:subspace approximation}
In the $k$-subspace problem, the input is a set $P$ of $n>d$ points in $\REAL^d$, the $d$-dimensional Euclidean space, along with an integer $k \in [1,d-1]$. The objective is to find a $k$-dimensional linear subspace ($k$-subspace, for short) $S\subseteq\REAL^d$ that closely represents these points according to a given loss function. This is a fundamental problem in mathematics and computational geometry with numerous applications; see surveys, e.g., in~\cite{kim2014robust,ta2014linear,zhang2014comparative,cao2016real,kanji2016self,bhutta2020smart,wendel2016self}.
A natural loss function is to compute the $k$-subspace $S$ that minimizes its sum over each Euclidean distance $\dist{p,S}:=\min_{s\in S}\norm{p-s}_2$from the $n$ input points, 
\begin{equation}\label{summ}
\sum_{p\in P}\mathrm{dist}^z(p,S)=\sum_{p\in P}\norm{p-XX^Tp}_2^z=\norm{ \left(\norm{p-XX^Tp}_2\right)_{p\in P}}_\zz,
\end{equation}
over every matrix $X\in\REAL^{d\times k}$ whose columns are mutually orthogonal unit vectors, i.e., $X^TX=I$, and $\Pi:=XX^T\in\REAL^{d\times d}$ is a projection matrix of rank $k$. 

For simplicity of notation, we assume that $P=(p_i)_{i=1}^n$ is an ordered set, although it may be arbitrary. By letting 
$A\in\REAL^{n\times d}$ denote the matrix whose $i$th row is $A_{i*}:=p_i$, the $k$-subspace median corresponds to a matrix $B$ that minimizes
\begin{equation}\label{ab}
\norm{A-B}_{2,\zz}:=\norm{ \left(\norm{A_{i*}-B_{i*}}_2\right)_{i=1}^n}_\zz,
\end{equation}
over every $k$-rank matrix $B\in\REAL^{n\times d}$, where $B_{i*}$ denote its $i$th row for $i\in\br{1,\cdots,n}$. In~\eqref{ab} we essentially want to approximate the general matrix $A$ by a matrix $B$ of the same size, but whose rank is only $k$. See Section~\ref{discussions} or, e.g.~\cite{song2017low,li2016tight}. This is a classic problem with a lot of  motivation in both theory and practice; see Section~\ref{why}.

However, as summarized in Section~\ref{related work section}, there are very few solvers for the $k$-subspace median problem. They are usually randomized, but, more importantly, their running time is exponential in $k$. This is impractical for modern applications, especially in the current era of data-driven AI science, where billions of parameters are used for a model, and not only the original dimension is high, but also the embedded one. 

In this paper, we thus focus on the $k$-subspace median ($\KSM$) problem, i.e., the mixed $\ell_{q,\zz}:=\ell_{2,1}$ norm, but expect it to be easily extended to any constants $q,z>1$. 

\paragraph{Why it is hard? }Since minimizing a quadratic function yields a linear function, PCA/SVD can use linear algebra for the case of $k$-subspace mean ($\zz=2$). For the case $k=d-1$, the problem is not linear but still convex, as we wish to minimize $\norm{Ax}_{1}$ over $\norm{x}_2=1$, which is the boundary of an Ellipsoid. Similarly, the $\ell_{z,z}$ norm, that is sometimes called $\ell_z$-norm, is convex also for $k<d-1$, due to the fact that we can switch between sum over rows to sum over columns~\cite{dasgupta2009sampling}. However, for the mixed-norm $\ell_{2,1}$ the problem is no longer convex for $k<d-1$, as the rank constraint is non-convex, which may explains the small number of its solvers, both in theory and practice. 

\paragraph{Data reduction. }There are many data reduction techniques that can be used to improve the running time of a given solver from polynomial to near-linear in the size of the input, including the one in this paper; see Section~\ref{datareduction}. This fact motivated our focus on polynomial, but possibly super-linear time approximation algorithm. However, to our knowledge, there are only a handful of such provable randomized algorithms or heuristics, and whose running time or approximation factors are usually exponential in $k$. In fact, most of the related coresets we found either has size exponential in $k$ or $d$ (e.g.~\cite{feldman2006coresets}), suited for $\ell_{\zz,\zz}$ (non-mixed) norm~\cite{dasgupta2009sampling}, and assume convexity of objective function~\cite{tukan2020coresets}.

The natural open problem that we answer affirmably in this paper is thus as follows.

\medskip
\paragraph{Is there a deterministic polynomial-time algorithm that computes the $k$-subspace median of a given set of points, up to an approximation factor that is not exponential in $k$?} \label{theory 1 q} \\

In particular, we aim to approximation factors that are smaller than $k$, even when $k\in \Omega(d)$.

\section{Related Work}
\label{related work section}
In this paper, we aim to deterministic algorithms for the case $\zz=1$ and non-fixed $k$, such as $k\in \Omega(d)$, but give more general results in this section.


\paragraph{Affine subspace (flat)}. A straightforward extension for the $k$-subspace median is the \emph{$k$-flat median} that minimizes the sum of distances among every \emph{affine} $k$-subspace of $\REAL^d$. However, there are simple reductions to the (non-affine) linear case; see e.g.~\cite{maalouf2020tight} and Section~\ref{discussions}. 

\paragraph{The case $k=0$.}
In this case, the only $0$-subspace is the origin, but the $0$-flat median is the median point, also called the \emph{Fermat-Weber Problem}~\cite{hansen1983recent}, which minimizes the sum of Euclidean distances to the $n$ inputs points. This is a convex problem that cannot be solved exactly~\cite{cockayne1969euclidean}, but a $\rbr{1+\eps}$-approximation can be computed in $O\rbr{nd\log^{3}{\frac{1}{\eps}}\log{\frac{1}{\delta}}}$ time with a probability of at least $1-\delta$~\cite{cohen2016geometric}.

\paragraph{The case $k=d-1$.}
In this case, the required subspace is a hyperplane, the orthogonal vector to this hyperplane is a vector, and the problem is convex since $\norm{Ax}_{2,1}$ is a mixed norm of $x$ which is a convex function. Hence, techniques such as the Ellipsoid method might give an $O(d)$-approximation in polynomial time.

\paragraph{The case $k\in\br{1,\ldots,d-2}$.}
Here, the $k$-subspace median problem is neither convex nor concave. We could not find any polynomial-time algorithm for the case where $k$ (and thus also $d>k$) is large; i.e., part of the input for the subspace-median problem. However, there are some randomized polynomial-time algorithms for the case where $k$ is constant. All of these algorithms are randomized and take an exponential time in the dimension $k$ of the desired subspace.

Shyamalkumar and Varadarajan~\cite{shyamalkumar2007efficient} suggested a randomized $\rbr{1+\eps}$-approximation for the $k$-subspace median that takes time
$O\rbr{\frac{ndk}{\eps}\ln{\frac{k}{\eps}}}=nd\rbr{\frac{k}{\eps}}^{O(1)}$, with a probability of success $\sfrac{1}{2^{O\rbr{\frac{k^2}{\eps}\ln^2{\frac{k}{\eps}}}}}=\sfrac{1}{2^{\rbr{\sfrac{k}{\eps}}^{O\rbr{1}}}}$.
Hence, to obtain a probability of success larger than, say, $\frac{1}{2}$, using amplification~\cite{hromkovivc2005success}, we need to run the algorithm number of times that is exponential in both $k$ and $\frac{1}{\eps}$. 

Deshpande and Varadarajan~\cite{deshpande2007sampling} refined this result, showing that it is possible in $nd\rbr{\frac{k}{\eps}}^{O\rbr{1}}$ time to produce a subset of $r=\rbr{\frac{k}{\eps}}^{O\rbr{1}}$ points, whose span contains a $\rbr{1+\eps}$-approximation to the $k$-subspace median. By projecting the $n$ input points onto the span of these $r$ points, one can find the $k$-subspace median in the time exponential in the smaller dimension $r$.

Feldman et al.~\cite{feldman2010coresets} suggested a randomized $\rbr{1+\eps}$ approximation that runs in
\begin{equation*}
    nd\rbr{\frac{k}{\eps}}^{O\rbr{1}}+\rbr{n+d}\exp\rbr{\rbr{\frac{k}{\eps}}^{O\rbr{1}}}.
\end{equation*}
where $
\exp\rbr{x}\coloneqq e^{x}$ by applying the algorithm of Shyamalkumar and Varadarajan on a coreset that will be defined later in this section.  Later, Clarkson and Woodruff~\cite{clarkson2015input} improved the running time to 
\begin{equation*}
    O\rbr{\nnz{P}}+\rbr{n+d}\rbr{\frac{k}{\eps}}^{O\rbr{1}}+\exp\rbr{\rbr{\frac{k}{\eps}}^{O\rbr{1}}},
\end{equation*}
where $\nnz{P}\leq nd$ is the number of non-zero entries in the input set.

\paragraph{$\ell_{\zz,\zz}$ norm.} 
As stated in~\eqref{minb}, the subspace median aims to compute the $k$-rank matrix $B$ that minimizes $\norm{A-B}_{2,1}$ for a given matrix $A\in\REAL^{n\times d}$. For the case $\norm{A-B}_{1,1}$, 
Markopoulos et al.~\cite{markopoulos2014optimal} showed an algorithm that minimizes the sum of $\ell_1$ norms from the points to the $k$-subspace, a $\sqrt{d}$-approximation to $\KSM$. However, their running time is exponential in the dimension $d$.
Randomized algorithms for polynomial time approximation for the $\ell_1$ norm were suggested in~\cite{chierichetti2017algorithms,song2017low}. 

For the case of minimizing $\norm{A-B}_{p,p}$, we refer to~\cite{dasgupta2009sampling} for polynomial time approximations in polynomial time. Other norms, such as Schatten and Spectral norms, were studied by Bakshi et al.~\cite{bakshi2022low}, and Fack and Kosaki~\cite{fack1986generalized}.

\paragraph{Software. }Although there are very few provable algorithms for the $k$-subspace median, we expected to find more software and heuristics, but it was not the case. Most of the software packages we found handle restricted version of the $\ell_{2,2}$ norm, such as regularization versions, or the $\ell_{1,1}$; see references in~\cite{markopoulos2017efficient}.
For example, we tried to run our experimental results with the new toolbox~\cite{markopoulos2025l1pca} for $\ell_{1,1}$ but their execution was not completed after few nights, unless either the number $n$ of points or dimension $k$ of the subspace were very small.

\section{Our Contribution}
\label{our contribution}
\label{our contribution:theory}
This work provides an affirmative answer to the open question in Section~\ref{background:subspace approximation}. That is, a deterministic algorithm that computes an approximation to the $k$-subspace median of a given set of points, whose running time and approximation factors are independent of both $n$ and $k$. More precisely, the approximation multiplicative factor is $\sqrt{d}$, which is sub-linear in $k$ even for $k=\Omega(d)$. The algorithm is fully polynomial in all its parameters $n,d$ and any $k\geq1$; see Corollary~\ref{maincol} for more precise details. 

The running time may be improved by applying the suggested algorithm on one of the appropriate coresets or sketches from the previous section to obtain a running time that is linear or near-linear in $n$. Although this would turn our deterministic algorithm into a random one, we are not aware of such a polynomial-time approximation in this case. Similarly, dimensionality reduction techniques from the previous section can enable replacing the approximation factor $\sqrt{d}$ by $k^{O(1)}$, but the factor would then be larger for values such as $k=\Omega(d)$. 

Finally, we expect that our simple technique would hold for other mixed $\ell_{2,\zz}$ norms such as the $k$-subspace center ($\zz=\infty$) but leave this for future work. 

\paragraph{Implementations and experiments. }As explained in the previous section, Unlike many implementations for PCA/SVD and their approximations, we could not find implementations for existing provable algorithms or experimental results for computing the $\KSM$ or even its variants. We thus implemented them together with our own algorithm and publish as open code~\cite{myrepo}. Experimental results that show a significance advantage in running time, approximation or both are provided in Section~\ref{experimental results}.

\paragraph{Novelty. }Existing related work to minimize mixed $\ell_{\zz,\zz}$ norms is heavily based on the fact that the two norms are the same, and thus we can change the sum of errors (entries in the distances matrix) between columns and rows; see e.g.~\cite{dasgupta2009sampling}. In this case, the problem is reduced to the convex one $\norm{Ax}_{\zz}$, as explained in the previous section. However, we could not see how to generalize this technique for $\ell_{2,\zz}$ where $\zz\neq 2$ (as in $k$-PCA/SVD).

\label{novelty}

\begin{figure*}[!hbt]
    \centering
    \includegraphics[width=0.8\linewidth]{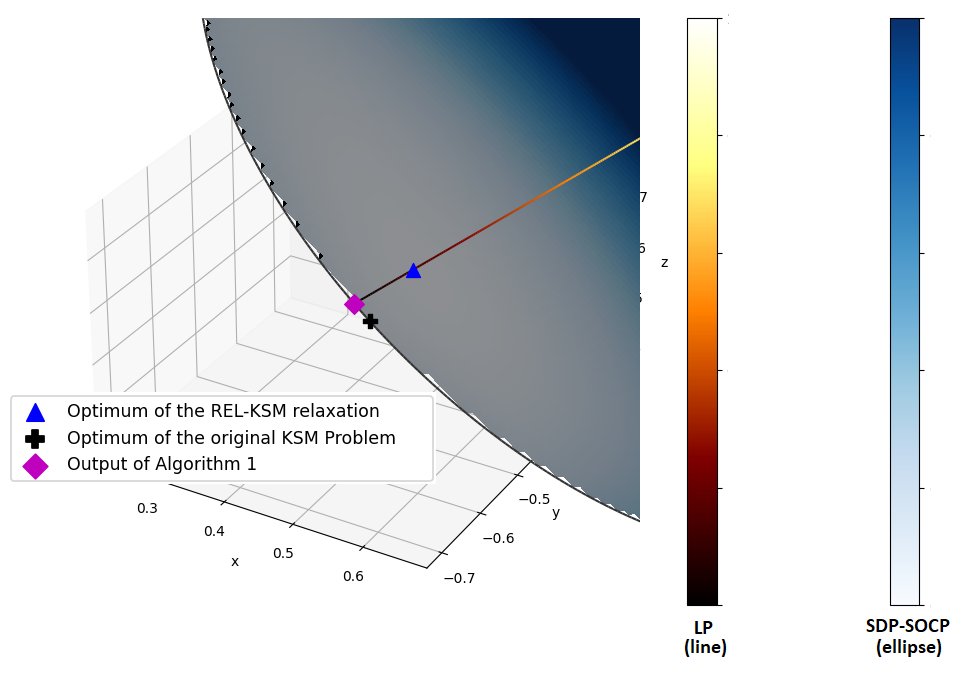}
    \caption{\scriptsize An illustration of the search space in Algorithm~\ref{alg:median_svd} for the case $d=2$ and $k=1$ (line-median in the plane). Every point is a matrix in $\REAL^{2\times 2}$. The gray ellipsoid consists of $\psd_2\subseteq\REAL^{2\times 2}$ which is the search space of Problem~\ref{not:median_socp_rel}, and its (non-convex) boundary consists of those matrices whose rank is $d-k=1$ (projection matrices) which is the search space of the original $\KSM$ problem~\eqref{not:median_orig}. The values of the function of the mixed SDP-SOCP optimization problem~\eqref{not:median_socp_rel} are manifested in the blue bar while the red bar manifests the cost function of the Linear Programming~\eqref{alg:linear_programming}. The projection matrix of the (optimal) $1$-subspace median is the black cross, $X^*\in \psd_2$ (blue triangle) minimizes the relaxed problem~\eqref{not:median_socp_rel}, and the pink square denote its ``closest" projection matrix $E$, which is the output of the algorithm.
     }
    \label{fig:algorithm_illustration}
\end{figure*}

To this end, we suggest a new optimization technique as shown in Fig.~\ref{fig:algorithm_illustration}. Firstly, unlike most related papers, we run the algorithm in the space of symmetric $d\times d$ matrices and not on the original space $\REAL^d$ of $d$-dimensional points (vectors). For example, in $d=2$ dimensions, we search over the space of symmetric matrices $\begin{pmatrix} x & y \\ y & z\end{pmatrix}$.
Secondly, we had to handle the hardness of the $k$-subspace median, which is related to the fact that the set of feasible solutions (projection matrices) is neither convex nor concave; see the boundaries of the ellipsoid in Fig.~\ref{fig:algorithm_illustration}.

To handle this issue, we solved the corresponding convex problem on the convex hull of the feasible set (the filled ellipsoid in Fig.~\ref{fig:algorithm_illustration}). Although the optimal solution is not on this boundary, we prove that a good approximation to the optimal solution is feasible; see blue $\bigtriangleup$ in Fig.~\ref{fig:algorithm_illustration}.

Finally, we show how to turn this infeasible solution into a feasible solution; see the blue $\bigtriangleup$ and the pink $\diamond$ in Fig.~\ref{fig:algorithm_illustration}, respectively. We achieved it by a careful design of a subspace of our matrix space that passes through the infeasible solution and whose intersection with the feasible set (ellipsoid) yields a solution that is still a provably good approximation but also feasible. In other words, we approximate a general PSD matrix by its projection to a closest $k$-rank matrix with respect to the sum of (non-squared) distances.

This non-trivial projection, whose details are described in Section~\ref{algorithm overview}, can be considered as a variant of the classic Eckart–Young–Mirsky~\cite{wang2015least} and the Courant–Fischer~\cite{ikebe1987monotonicity} theorems for approximating a matrix by a $k$-rank matrix, where the Frobenius norm is replaced by the $\ell_{2,1}$ mixed norm, and instead of selecting the largest $k$ singular values, we use a more involved technique. While we use the standard central path method in Section~\ref{central} to solve our SDP problem on the second-order cone, another challenge was to compute an initialization matrix which is provably close to the optimum in Lemma~\ref{complexity:first iteration lemma}.


We expect that this technique should work for other mixed norms and convex functions over $k$-rank matrices.


\section{Notation and Definitions}
\label{notations}
Throughout this work, we assume that we are given integers $d\geq2$ and $n\geq d$ that represent the dimension and cardinality of the input set of points, respectively. We define $\REAL^d$ as the set of real $d$-dimensional column vectors, where a single column is denoted by $x:=(x_1,\cdots,x_d)\in\REAL^d$. The set of all $n\times d$ real matrices is $\REAL^{n\times d}$. The determinant of a matrix $X\in\REAL^{n\times n}$ is $\abs{X}\coloneqq\det{X}$. The rank and trace of a matrix $A$ are denoted by $\rank{A}$ and $\trace{A}$, respectively. Furthermore, we define $\cbr{n}\coloneqq\br{1,\ldots,n}$, $\REAL_+\coloneqq\br{x\in\REAL\middle|x\geq0}$, $\REAL_{++}\coloneqq\br{x\in\REAL\middle|x>0}$. For a set $\br{k_i}_{i=1}^n$ of positive integers and $v_i\in\REAL^{k_1}$, we define $v\coloneqq\rbr{v_1\mid \ldots\mid v_n}\in\REAL^{\sum_{i=1}^nk_i}$ as the column vector of their concatenation. For a vector $v\in\REAL^n$, we define $\diag{v}\in\REAL^{n\times n}$ as the diagonal matrix whose diagonal is $v$, and for a matrix $X\in\REAL^{n\times n}$ we denote by $\diag{X}\in\REAL^n$ the vector of the values on the diagonal of $X$.

The running time it takes to invert a matrix in $\REAL^{n\times n}$ is denoted by $O(n^{\omega})$.

The \emph{positive semi-definite (PSD) cone} is denoted by
\begin{equation}\label{psd}
\psd_d\coloneqq\br{X\in\REAL^{d\times d}\middle|X^T=X \text{ and } \forall v\in\REAL^d, v^TXv\geq0},
\end{equation}
its interior is~\cite{boyd2004convex}
\[
\ipsd_d\coloneqq\br{X\in\REAL^{d\times d}\middle|X^T=X \text{ and } \forall v\in\REAL^d, v^TXv>0}.
\]
The \emph{second order cone} is denoted by 
\begin{equation}\label{soc}
\soc_d:=\br{\rbr{v,t}\in\REAL^{d}\times [0,\infty)\mid\norm{v}_2\leq t},
\end{equation}
and its interior is~\cite{boyd2004convex}
\[
\isoc_d:=\br{\rbr{v,t}\in\REAL^{d}\times [0,\infty)\mid\norm{v}_2< t},
\]

We define $\doubleOne_n:=(1,\cdots,1)$ and $0_n:=(0,\cdots,0)$ to be the $n$ dimensional column vectors of ones and zeros, respectively, and $0_{n\times n}$ to be the $n\times n$ zero matrix. The $d\times d$ identity matrix is denoted by $I_d$, or $I$ if $d$ is clear from the context. For every pair $X,Y\in \psd_d$, we define the partial order
\begin{equation*}
    X\preceq Y \iff Y-X\in \psd.
\end{equation*}

The \emph{eigenvalue decomposition} $X=VDV^T$ of a matrix $X\in\mathbb{S}^+_n$ satisfies $V^TV=I$, where $D\in \mathbb{S}_n^+$ is a diagonal matrix. It exists for every matrix $X\in\mathbb{S}^+_n$~\cite{golub1996matrix}. If $D\in\br{0,1}^{n\times n}$ is a binary diagonal matrix, i.e., it has $k\geq 0$ diagonal entries of ones and $(d-k)$ zeros, then $X$ is called a $k$-rank \emph{projection matrix} and satisfies $X^2=X$.
Its \emph{corresponding $k$-subspace} is the column space of $X$ (or $V$).

The \emph{$k$-median subspace} (or $\KSM$, for short) of a set $P=(p_i)_{i=1}^n$ of $n$ points in $\REAL^d$, minimizes the sum of Euclidean distances to its points. 
\begin{align}
    \label{not:median_orig}
    \min_{X\in\REAL^{d\times d}}  \quad &  \sum_{i=1}^n\norm{Xp_i}_2
    \\\nonumber
    s.t. \quad & X^2=X^T=X
    \\\nonumber
    & \rank{X}=d-k.
\end{align}
We denote the optimal solution by $\ksm(P,k)$. More formally, 
\begin{equation*}
    \ksm:\Big( (\REAL^d)^n\times\cbr{d-1}\Big)\rightarrow\REAL_+
\end{equation*}
is the function that maps every $2$-tuple $\rbr{(p_i)_{i=1}^n,k}\in(\REAL^d)^n\times\cbr{d-1}$ to the sum of Euclidean distances to a $k$-subspace median of the ordered set $(p_i)_{i=1}^n$ of points.

For a given $\alpha\geq1$, a $k$-subspace $\sub\subseteq\REAL^d$ is an \emph{$\alpha$-approximation} to the $\KSM$ of the set $P$, if its sum of Euclidean distances to the points is the same as $\ksm\rbr{\rbr{p_i}_{i=1}^n,k}$, up to a multiplicative factor of $\alpha$.
Formally, 
\begin{align*}
    \sum_{i=1}^n\dist{p_i,S}\leq\alpha\ksm\rbr{P,k}.
\end{align*}

\section{Main Algorithm}
\label{algorithm}
This section presents our main approximation algorithm. It gets a set $P$ of points in $\REAL^d$ with an integer $k\in[d-1]$,  and computes a projection matrix of rank $(d-k)$ whose null space is a $k$-subspace in $\REAL^d$. We will prove that it is an $\sqrt{d}$-approximation to $\ksm\rbr{P,k}$, i.e.,  minimizes the sum of Euclidean distances to the points of $P$ up to a multiplicative factor of $\sqrt{d}$ as defined in~\eqref{not:median_orig}.

The input of the algorithm is the input to the $\KSM$ problem: a set $P=(p_i)_{i=1}^n$ of $n$ points in $\REAL^d$, together with an integer $k\in [d-1]$.
The main technical result is a relaxation of the $\KSM$ problem to the following mixed SDP-SOCP minimization problem, with the same input $P$ and $k$:
\begin{equation}
\begin{split}
    \label{not:median_socp_rel}
    \min_{\substack{X\in\REAL^{d\times d} \\ y\in\REAL^n}} \quad &  \doubleOne_n^Ty
    \\
    s.t. \quad & X^T=X
    \\
    & \trace{X}=d-k
    \\
    & 0_{d\times d} \preceq X \preceq I_d
    \\
    & \norm{Xp_i}_2\leq y_i \quad \forall i\in\cbr{n}.
\end{split}
\end{equation}

The optimal solution of this problem is denoted by $\rlksm(P,k)$. That is,  the function
\begin{equation}\label{rlksm}
    \rlksm:(\REAL^d)^n\times\cbr{d-1}\rightarrow\REAL_+,
\end{equation}
maps every $2$ tuple $\rbr{P,k}\in(\REAL^d)^n\times \cbr{d-1}$ to the minimizers of Problem~\eqref{not:median_socp_rel}. 

%
\begin{algorithm}[!hbt]
\DontPrintSemicolon
\caption{$\ksma\rbr{P,k}$; see Lemma~\ref{alg:correctness theorem}.}
\label{alg:median_svd}
    \SetKwInOut{Input}{Input}
    \SetKwInOut{Output}{Output}
    \Input{A set $P$ of $n$ points in $\REAL^d$, and an integer $k\geq1$.}
    \Output{A $k$-dimensional linear subspace $S\subseteq \REAL^d$} 
    Compute a minimizer $\rbr{X^*,y^*}\in \psd_d\times\REAL^n$ of the convex problem~\eqref{not:median_socp_rel}. \\ \tcp{See Proposition~\ref{complexity:ksm central path iterations bound} and Lemma~\ref{complexity:first iteration lemma}.} \label{alg:socp-sdp line}
    Compute a spectral decomposition $X^*=VDV^T$ of $X^*$. \label{alg:svd line} \BlankLine\tcp{i.e., $D$ is diagonal and $V^TV=I_d$.}
    \For {every $i\in\cbr{n}$ \label{alg:subspace start line}}{Set $v^{\rbr{i}} \gets Vp_i$\label{vline}}
        \label{qline}Let $q\in[0,\infty)^d$ be the vector whose $j$th entry is $q_j:=\sum_{i=1}^n\abs{v_j^{\rbr{i}}}$, for every $j\in[d]$.\\
        \tcp{$v_j^{\rbr{i}}$ is the $j$th entry of $v^{(i)}$}
        \label{alg:subspace end line}
    Set $\zeta \gets 0_d$ \;\label{alg:optimum start line}
    Let $\ind\subseteq\cbr{d}$ denote the indices of the $d-k$ smallest entries of $q$; ties broken arbitrarily.\;\label{alg:cost start line}
    Set $\zeta\in \br{0,1}^d$ such that $\zeta_i\gets1$ if and only if $i\in \ind$\;\label{alg:cost end line}
    Set $E\gets V^T\diag{\zeta}V$ \;\label{alg:optimum end line}
    Let $S\subseteq\REAL^d$ denote the corresponding $k$-subspace of the projection matrix $V(I-E)V^T$.\\ \tcp{$S$ is the column space of $(I-E)V$.}\label{alg:s definition}
    \Return $S$ \;\label{alg:return line}
\end{algorithm}

\subsection{Algorithm overview.}
\label{algorithm overview}

To demonstrate the intuition behind Algorithm~\ref{alg:median_svd}, consider the simplest case where the input points are on the plane $(d=2)$ and the subspace median is a line through the origin ($k=1$). 
Each 3-dimensional point $\rbr{x,y,z}$ in Fig.~\ref{fig:algorithm_illustration} represents a symmetric matrix $\begin{pmatrix} x & y \\ y & z\end{pmatrix}$ in $\REAL^{2\times 2}$. The search space of Algorithm~\ref{alg:median_svd} is the union over every such possible matrix, and is denoted by the gray ellipsoid in Fig.~\ref{fig:algorithm_illustration}. In contrast, each point on the ellipsoid's boundary consists of a $(d-k)$-rank projection matrix, which is the search space of the original $\KSM$ problem in~\eqref{not:median_orig}. The hardness of solving the $\KSM$ problem arises from the non-convexity of the feasible set in Problem~\eqref{not:median_orig}, which is the set of $\rbr{d-k}$-rank matrices which corresponds to the non-convex boundary of the (convex) ellipsoid, which is an empty ellipsoid.

Instead of solving the NP-hard $\KSM$ problem in~\eqref{not:median_orig} directly, and compute the optimal point in the empty (non-convex) ellipsoid, which is denoted by a black cross in Fig.~\ref{fig:algorithm_illustration}, we compute the optimal solution $X^*$ over the (full, convex) ellipsoid, which is denoted by a blue triangle in the figure. This is the minimizer of Problem~\eqref{not:median_socp_rel}, which is solved in Line~\ref{alg:socp-sdp line} of Algorithm \ref{alg:median_svd}. It is a convex mixed SDP-SOCP relaxation of Problem~\eqref{not:median_orig}, and returns a 3-dimensional (more generally, $\frac{d\rbr{d+1}}{2}$-dimensional) point $X^*\in \mathbb{S}^+_2$ in the space of positive semi-definite $2\times 2$ matrices inside the ellipsoid (the blue triangle).

Although this minimizer (matrix) $X^*$ of Problem~\eqref{not:median_socp_rel} approximates the objective function of the original problem~\eqref{not:median_orig}, it is not a feasible solution. That is, the resulting matrix is inside the ellipsoid and thus in $\mathbb{S}^+_2$, but not necessarily a $(d-k)$-rank matrix, i.e., a projection matrix, as it is not on the ellipsoid boundary. To turn it into a projective matrix, we project it on the boundary of the ellipsoid. This is the pink triangle in Fig.~\ref{fig:algorithm_illustration}, which is the projection of the blue triangle onto the empty ellipsoid. 

In the algorithm, this projection is computed in Line~\ref{alg:svd line}, where the eigendecomposition $X^*=V^TDV$ is computed. Since $X^*$ is PSD, its eigenvalues along the diagonal of $D$ are non-negative, while a projection matrix has only binary eigenvalues. In Lines~\ref{alg:subspace start line}-\ref{alg:optimum end line}, we compute the ``closest" $(d-k)$ projection matrix to $X^*$ in some sense. We use a careful ``rounding" of each entry in $D$ to $1$ or $0$. The resulting diagonal matrix is $\diag{\zeta}\in \br{0,1}^{2\times 2}$, and Line~\ref{alg:cost start line} insures that it will consist of only $d-k$ entries of ``1". 

After rounding the entries of $\diag{D}$ to the binary entries of $\zeta$, the new matrix $E:=V^T\diag{\zeta}V$ would be a projection matrix that corresponds to a $k$-subspace which is spanned by a subset of $k$ columns from $V$. However, it will no longer minimize Problem~\eqref{not:median_socp_rel}. It is left to choose a rounding procedure that would provably bound the approximation factor of $E$ with respect to $X^*$.

Without loss of generality, assume that $V=I_d$, otherwise rotate the coordinate system. In this case, the non-zero entries of $\zeta$ correspond to a subset of $k$ among the $d$ axes, which would span the output $k$-subspace. In Line~\ref{vline}, we compute the projection $v^{(i)}_j$ of the $i$th input point on the $j$th axis, and in Line~\ref{qline} $q_j$ is the sum of these projections. Geometrically, $q_j$ is the sum of
distances of the input points to the hyperplane that is orthogonal to the $j$th axis. This is in contrast to the matrix $D$, whose diagonal entries correspond to the sum of \emph{squared} distances to each such hyperplane. Do not confuse $q_j$ with $\sqrt{D_j}$ which is the squared root of the sum of squared distances to these hyperplanes. 
Since the $k$-subspace median minimizes its sum of distances to the points, it makes sense that in Line~\ref{alg:cost start line}, we choose the output subspace to be the one that is spanned by the $k$-axes which minimizes this sum of distances.

\section{Proof of correctness}
The goal of this section is to prove the main technical result in Lemma~\ref{alg:correctness theorem}. That is, that Algorithm~\ref{alg:median_svd} indeed returns a $\sqrt{d}$ approximation to the $\ksm$ of the input set $P$, and that its running time is polynomial in the input.
The proof is based on the intuition of Section~\ref{algorithm overview}.

Consider the value of $\zeta$ that is computed during the execution of Line~\ref{alg:optimum end line}. The next proposition states that $\zeta$ minimizes the linear program that is defined in \eqref{alg:linear_programming}. That is, taking the largest $d-k$ entries of $q$, is an ``binary rounding" that yields the largest value $\lambda^Tq$ over any combination $\lambda\in[0,1]^d$ whose sum is $\norm{\lambda}_1=d-k$. The proof is a straight-forward application of the ``Fundamental theorem of linear programming"~\cite{tardella2011fundamental} and can be considered as a variant of the Eckart–Young–Mirsky~\cite{wang2015least} and the Courant–Fischer~\cite{ikebe1987monotonicity} theorems for the case of $\zz=2$, as in $k$-SVD. A main difference is that these theorems take the largest $k$ values from the same ordered list of singular values, regardless of $k$. In our case, every value of $k$ yields a different set of candidate entries in $\lambda$. This is probably related to the monotonicity property of the $k$-subspace mean which is contained in the $(k+1)$-subspace mean, unlike the $k$-subspace median case. Without loss of generality, the proposition below assumes that the entries of $q$ are sorted. Otherwise, we sort them and rearrange them back after computing $\lambda$.

As a result, $E=V^T\diag{\zeta}V$ is not only a $d-k$ projection matrix that lies on the boundary of the gray ellipsoid of Fig.~\ref{fig:algorithm_illustration}, but also a minimizer of $\lambda^Tq$ above. This fact will be used in our main proof to bound the approximation factor. 
\begin{proposition}
\label{alg:lp_proposition}
Let $q\in[0,\infty)^d$ be a vector that satisfies $q_1\leq \cdots \leq q_d$. Let $k\in[d-1]$ be an integer, and $\zeta=(1,1,\cdots,0,0)\in \br{0,1}^d$ be a binary vector that satisfies $\zeta_i=1$ if and only if $i\in [d-k]$. Then
\begin{equation}
\label{alg:linear_programming}
    \zeta\in\arg\min_{\lambda}\lambda^Tq,
\end{equation}
where the minimum is over every $\lambda\in[0,1]^d$ whose sum of entries is $\sum_{i=1}^d \lambda_i=d-k$.
%
\end{proposition}
\begin{proof}
Minimizing the linear function $\lambda^Tq$ over the polytope $\br{\lambda\in[0,1]^d\mid\sum_{i=1}^d \lambda_i=d-k}$ is a linear program. The proposition then follows from the fundamental theorem of linear programming, see e.g.~\cite{tardella2011fundamental}.
\end{proof}

Using Proposition~\ref{alg:lp_proposition}, we can prove that the output of Algorithm~\ref{alg:median_svd} is a $\sqrt{d}$-approximation as follows.

\begin{theorem}[Correctness of the Algorithm~\ref{alg:median_svd}]
\label{alg:correctness theorem}
Let $k\in [d-1]$ be an integer and $P=(p_i)_{i=1}^n\in (\REAL^d)^n$. Let $S\subseteq\REAL^{d\times d}$ be the output of a call to $\ksma\rbr{P,k}$; see Algorithm~\ref{alg:median_svd}. Then $S$ is a $\sqrt{d}$-approximation to the $k$-subspace median of $P$. That is,
\begin{equation*}
\sum_{i=1}^n\dist{p_i,S}\leq\sqrt{d}\cdot \ksm(P,k).
\end{equation*}
\end{theorem}
\begin{proof}
Consider the values of $X^*$, $y^*$, $V$, $D$, $\br{v^{\rbr{i}}}_{i=1}^n$, $q$, $\zeta$, and $E$ during the execution of Line~\ref{alg:return line} of Algorithm \ref{alg:median_svd}.
Since $\zeta$ consists of $d-k$ ones and $k$ zeros, $E$ is a symmetric matrix as
\begin{align}
    \nonumber
    E^2=\rbr{V^T\diag{\zeta}V}^2=V^T\diag{\zeta}^2V=V^T\diag{\zeta}V=E,
\end{align}
and have rank $k$, as
\begin{align}
    \nonumber
    \rank{E}=\rank{V^T\diag{\zeta}V}=\rank{\diag{\zeta}}=d-k.
\end{align}
Hence, $E$ is a projection matrix that corresponds to a $k$-subspace in $\REAL^d$. It is left to prove that this feasible solution to Problem~\eqref{not:median_orig} indeed approximates the $k$-subspace median.

Indeed, by~\eqref{not:median_socp_rel} $x^TX^*x\in [0,1]$ for every unit vector $x\in\REAL^d$, and thus its eigenvalues are in $[0,1]$. That is, $D\in[0,1]^{d\times d}$. We also have $$\norm{\diag{D}}_1=\trace{D}=\trace{V^TDV}=\trace{X^*}=d-k,$$ where the last equality holds by~\eqref{not:median_socp_rel}. We can thus substitute $\lambda:=\diag{D}$ in Proposition~\ref{alg:lp_proposition} to obtain
\begin{equation}
    \label{alg:lp_otimum}
    \zeta^Tq\leq\diag{D}^Tq.
\end{equation}

For every $\lambda\in\REAL^d_+$, 
\begin{align}
    \label{alg:correctness:affine subspace 1 equality}
    \lambda^Tq&=\sum_{j=1}^d\lambda_j\sum_{i=1}^n\abs{v^{\rbr{i}}_j}
    \\\label{alg:correctness:affine subspace 2 equality}
    &=\sum_{i=1}^n\sum_{j=1}^d\abs{\lambda_jv^{\rbr{i}}_j}
    \\\label{alg:correctness:affine subspace 3 equality}
    &=\sum_{i=1}^n\norm{\diag{\lambda}v^{\rbr{i}}}_1
    \\\label{alg:correctness:affine subspace 4 equality}
    &=\sum_{i=1}^n\norm{\diag{\lambda}Vp_i}_1
\end{align}
where \eqref{alg:correctness:affine subspace 1 equality} and \eqref{alg:correctness:affine subspace 4 equality} are by the definition of $\br{v^{\rbr{i}}}_{i=1}^n$ and $q$ respectively,~\eqref{alg:correctness:affine subspace 2 equality} holds because $\lambda\in\REAL^d_+$, and \eqref{alg:correctness:affine subspace 3 equality} holds by the definition of $\norm{\cdot}_1$ and because $\diag{\lambda}$ is diagonal. Furthermore,
\begin{align}
    \label{alg:correctness:lp optimum 1 equality}
    \sum_{i=1}^n\norm{\diag{\zeta}Vp_i}_1&=\zeta^Tq
    \\\label{alg:correctness:lp optimum inequality}
    &\leq \diag{D}^Tq
    \\\label{alg:correctness:lp optimum 2 equality}
    &=\sum_{i=1}^n\norm{DVp_i}_1.
\end{align}
where \eqref{alg:correctness:lp optimum 1 equality} and \eqref{alg:correctness:lp optimum 2 equality} hold by substituting, respectively, $\lambda:=\zeta$ and $\lambda:=\diag{D}$ in \eqref{alg:correctness:affine subspace 4 equality}, and \eqref{alg:correctness:lp optimum inequality} holds by \eqref{alg:lp_otimum}.
Therefore,
\begin{align}
    \label{alg:correctness:cauchy-schwarz bound 1 inequality}
    \sum_{i=1}^n\norm{\diag{\zeta}Vp_i}_2
    &\leq\sum_{i=1}^n\norm{\diag{\zeta}Vp_i}_1
    \\\label{alg:correctness:cauchy-schwarz bound 2 inequality}
    &\leq\sum_{i=1}^n\norm{DVp_i}_1
    \\\label{alg:correctness:cauchy-schwarz bound 3 inequality}
    &\leq\sqrt{d}\sum_{i=1}^n\norm{DVp_i}_2,
\end{align}
where \eqref{alg:correctness:cauchy-schwarz bound 2 inequality} holds by \eqref{alg:correctness:lp optimum 2 equality}, and \eqref{alg:correctness:cauchy-schwarz bound 1 inequality} and \eqref{alg:correctness:cauchy-schwarz bound 3 inequality} holds since $\sqrt{d}\norm{y}_1\leq \norm{y}_2\leq \norm{y}_1$ for every $y\in\REAL^d$.

Let $X\in \psd_d$ be a projection matrix that corresponds to the $k$-subspace median of $P$. That is, 
\[
\ksm\rbr{P,k}=\sum_{i=1}^n\norm{Xp_i}_2.
\]
By its definition in Section~\ref{notations}, a $(d-k)$ projection matrix in $\REAL^d$ has eigenvalues $0$ and $1$ with multiplicity $k$ and $d-k$, respectively. Therefore, its sum of eigenvalues is $\trace{X}=d-k$. Letting $y\in\REAL^n$ be the vector whose $i$th coordinate is $y_i:=\norm{Xp_i}_2$ for every $i\in[n]$, yields a feasible solution $\rbr{X,y}$ for Problem~\eqref{not:median_socp_rel}. Hence,
\begin{equation}\label{rlkbound}
  \rlksm(P,k)\leq \doubleOne_n^Ty=\sum_{i=1}^n\norm{Xp_i}_2=\ksm\rbr{P,k}.
\end{equation}
It follows that
\begin{align}
\label{SS}
\sum_{i=1}^n\dist{p_i,S}&=  \sum_{i=1}^n\norm{p_i-(I-E)Vp_i}_2=\sum_{i=1}^n\norm{Ep_i}_2\\
    \label{alg:correctness:final approximation 1 equality}
    &=\sum_{i=1}^n\norm{V^T\diag{\zeta}Vp_i}_2
    \\\label{alg:correctness:final approximation 2 equality}
    &=\sum_{i=1}^n\norm{\diag{\zeta}Vp_i}_2
    \\\label{alg:correctness:final approximation 1 inequality}
    &\leq\sqrt{d}\sum_{i=1}^n\norm{DVp_i}_2
    \\\label{alg:correctness:final approximation 3 equality}
    &=\sqrt{d}\sum_{i=1}^n\norm{V^TDVp_i}_2=\sqrt{d}\sum_{i=1}^n\norm{X^*p_i}_2
    \\\label{alg:correctness:final approximation 4 equality}
    &=\sqrt{d}\cdot\rlksm\rbr{P,k}
    \\\label{alg:correctness:final approximation 2 inequality}
    &\leq\sqrt{d}\cdot \ksm\rbr{P,k},
\end{align}
where~\eqref{SS} follows from the definition of $S$ in Line~\ref{alg:s definition}, \eqref{alg:correctness:final approximation 1 equality} is by the definition of $E$ in Line~\ref{alg:optimum end line}, \eqref{alg:correctness:final approximation 2 equality} and \eqref{alg:correctness:final approximation 3 equality} hold because $V$ is an orthogonal matrix, \eqref{alg:correctness:final approximation 1 inequality} is by \eqref{alg:correctness:cauchy-schwarz bound 3 inequality}, \eqref{alg:correctness:final approximation 4 equality} holds because $X^*=V^TDV$ is an optimal solution of Problem \eqref{not:median_socp_rel} by Line~\ref{alg:socp-sdp line}, and \eqref{alg:correctness:final approximation 2 inequality} is by~\eqref{rlkbound}. Therefore, $S$ is a $\sqrt{d}$ approximation to the $\KSM$ of $P$.
\end{proof}


%
%

\section{Running time}
\label{complexity analysis}
Lemma~\ref{alg:correctness theorem} states the correctness of our main algorithm, but not its running time.
In this section, we establish an upper bound for the running time of Algorithm~\ref{alg:median_svd}, which is $\rbr{ndk}^{O\rbr{1}}$, that is, polynomial in all the parameters; for further details, refer to Theorem~\ref{running}. As explained in the first sections, this running time can be reduced to near-linear in the input by paying additional small constant factor, by applying our algorithm on reduced (small) core sets. 

An essential technique for our analysis is the \emph{central path method}~\cite{nesterov1994interior}, explained in Section~\ref{complexity:central path method}. Subsequently, we apply this method to tackle our specific $\KSM$ problem in Section~\ref{complexity:k subspace medain}.

\label{complexity:k subspace medain}
The $\KSM$ Problem~\eqref{not:median_socp_rel} gets as input a set $P=\rbr{p_1,\cdots,p_n}$ of $n$ points in $\REAL^d$ and an integer $k\in[d-1]$. We reduce it to an input instance $G:=(f_0,F,A,b)$ for a self-concordance conic optimization problem, which is defined in Definition~\ref{selfproblam}. 
The variables to minimize in the $\KSM$ Problem~\eqref{not:median_socp_rel} are $X\in\REAL^{d\times d}$ and $y\in\REAL^n$. Although these domains are not subsets of $\REAL^n$ as in this definition, they are isomorphic to an Euclidean space in a sense that enables us to use the central path in Theorem~\ref{thm:cent}. This straightforward generalization of convex optimization from Euclidean space to Cartesian products of metric spaces is well known and is explained in text books such as~\cite{boyd2004convex,ben2001lectures,beck2017first}.

\begin{lemma}\label{xylem}
Let $P:=\rbr{p_i}_{i=1}^n$ be a set of $n\geq1$ points in $\REAL^d$. 
There is a conic instance $G(P):=(f_0,F,A,b)$ for the optimization problem~\eqref{not:median_socp_rel} whose $t$-relaxation, for every $t>0$, is the function $G_{t}:\dd(G)\to\REAL$ that maps every $(X,y)\in\dd(G)$ to 
\begin{equation}\label{gtt}
G_{t}(X,y):=t\doubleOne_n^Ty-\sum_{i=1}^n\ln\rbr{{y^2}_i-\norm{Xp_i}_2^2}-\ln\abs{X}
    -\ln\abs{I_d-X}.
\end{equation}
Moreover, $\deg(F)=2(n+d)$.
\end{lemma}
\begin{proof}
Let $(P,k)$ denote the input of Problem~\eqref{not:median_socp_rel}.
That is, $k\in [d-1]$ is an integer and $P:=\br{p_1,\cdots,p_n}$ is a set of $n$ points in $\REAL^d$.
The function $f_0$ and the set $F:=\br{f_i,K_i,\psi_i}_{i=1}^{n+2}$ of $m:=n+2$ logarithmic barriers are defined as follows:
\begin{itemize}
    \item $f_0:\left(\REAL^{d\times d}\times\REAL^n\right)\rightarrow\REAL$ is defined by $f_0\rbr{X,y}:=\doubleOne^Ty$.
    \item $f_1:\left(\REAL^{d\times d}\times\REAL^n\right)\rightarrow\REAL^{d\times d}$ is defined by $f_1\rbr{X,y}:=-X$.
    \item $f_2:\left(\REAL^{d\times d}\times\REAL^n\right)\rightarrow\REAL^{d\times d}$ is defined by  $f_2\rbr{X,y}:=X-I_d$.
    \item $f_{i+2}:\left(\REAL^{d\times d}\times\REAL^n\right)\rightarrow \left(\REAL^d\times\REAL\right)$ is defined by  $f_{i+2}\rbr{X,y}:=-\rbr{Xp_i,y_i}$, for every $i\in\cbr{n}$.
    \item $K_1:=K_2:=\psd_d$ is the positive semi-definite cone (PSD) in $\REAL^{d\times d}$; see~\eqref{psd}.
    \item $K_{i+2}:=\soc_d$, for every $i\in[n]$, is the second order cone (SOC) in $\REAL^d\times \REAL$; see~\eqref{soc}.
    \item $\psi_1:=\psi_2:=\psi_{\ipsd}$ as the generalized logarithm $\psi_{\ipsd}:\ipsd_d\rightarrow\REAL$ that maps every positive definite matrix $X\in \ipsd_d$ to $\psi_{\ipsd}(X):=\ln \abs{X}$, the determinant of its natural logarithm.
    \item $\psi_i+2:=\psi_{\isoc}$, for every $i\in[n]$, as the generalized logarithm $\psi_{\isoc}:\isoc_d\rightarrow\REAL$ 
    that maps every $(v,t)$ that satisfies $\norm{v}_2<t$ to $\ln (t^2-\norm{v}_2^2)\in\REAL$. 
\end{itemize}

The constraint $-f_1(X,y)\in  \interior{K_1}$ implies $X=-f_1\rbr{X,y}\in  \interior{K_1}=\ipsd_d$, i.e., $X\succ 0$.
Similarly, the constraint $-f_2(X,y)\in  \interior{K_2}$ implies $I_d-X=-f_2\rbr{X,y}\in  \interior{K_1}=\ipsd_d$, i.e., $X\prec I_d$. Combining these two constraints yields
\begin{equation}\label{f11}
-f_1(X,y)\times (-f_2(X,y))\in  \interior{K_1}\times \interior{K_2} \Leftrightarrow 0_{d\times d}\prec X \prec I_d.
\end{equation}
The constraint $-f_{i+2}(X,y)\in  \interior{K_i}$ implies $(Xp_i,y_i)=-f_{i+2}\rbr{X,y}\in  \interior{K_{i+2}}=\isoc_d$, i.e., 
\begin{equation}\label{fii}
-f_{i+2}(X,y)\in  \interior{K_i} \Leftrightarrow \norm{Xp_i}<y_i,
\end{equation}
for every $i\in [n]$. 

To obtain a conic instance $G$ for Problem~\eqref{not:median_socp_rel}, it is left to add the constraints $X^T=X$ and $\trace{X}=d-k$. These are linear constraints that correspond to the matrix $A$ and the vector $b$ in the Euclidean space, or the variables $(X,y)\in \REAL^{d\times d}\times \REAL^n$ in our case. The first two linear equality constraints on the variable $\rbr{X,y}$ in~\eqref{not:median_socp_rel} are expressed as $\trace{X}=d-k$ and $X-X^T=0$. These are affine linear matrix equality constraints. Indeed, let $b_1=d-k$, $b_2=0$, and $A_1,A_2:\REAL^{d\times d}\times\REAL$ be the linear functions that maps $(X,y)$ to  $A_1(X,y):=\sum_{i=1}^d X_{i,i}$, and $A_2(X,y):=X-\sum_{i,j\in[d]} X_{j,i}E^{i,j}$. Here, $X_{i,j}$ is the $(i,j)$th entry of $X$ in the $i$th row and $j$th columns, and $E^{i,j}\in\br{0,1}^{d\times d}$ is zero in every entry except $E_{i,j}:=1$.
The constraints $A_i(X,y)=b_i$ for $i\in\br{1,2}$ imply the remaining constraints of Problem~\eqref{not:median_socp_rel}
\begin{equation}\label{Ab}
A(X,y)=b \Leftrightarrow \trace{X}=d-k \text{ and } X-X^T=0.
\end{equation}

Combining~\eqref{f11},~\eqref{fii} and~\eqref{Ab}, and substituting $x:=(X,y)$ in Definition~\ref{relax} yields that the resulting domain of this conic optimization problem is
\begin{equation}\label{dd}
\begin{split}
\dd(G):&=
\bigcap_{i\in [n+2]} \br{(X,y)\in \REAL^{d\times d}\times\REAL^n \mid -f_i(X,y)\in \interior{K_i}, A_1(X,y)=b_1, A_2(X,y)=b_2}\\
&=\bigcap_{i\in [n+2]} \br{(X,y)\in \REAL^{d\times d}\times\REAL^n \mid  \norm{Xp_i}<y_i, 0_{d\times d}\prec X \prec I_d, \trace{X}=d-k, X=X^T}.
\end{split}
\end{equation}

Let $$(X,y)\in \arg\min_{(X,y)\in \dd(G)} f_0(X,y)=\arg\min_{\rbr{X,y}\in \dd(G)} \doubleOne^Ty.$$ By~\eqref{dd} $(X,y)\in \dd(G)$ implies
$X=X^T$, $\trace{X}=d-k$, $0_{d\times d}\prec X \prec I_d$, and $\norm{Xp_i}<y_i$ for every $i\in[n]$. 
Therefore $(X,y)$ is an optimal solution to Problem~\ref{not:median_socp_rel}, which proved the first claim in Lemma~\ref{xylem}.

The degree of $\psi_{\psd}$ is $d$, and the degree of $\psi_{\soc}$ is $2$, as proved e.g. in~\cite{boyd2004convex}, along with the proofs that every $(f,K,\psi)\in F$ is indeed a logarithmic barrier. Hence, 
\[
\deg(F)=2d+2n. 
\]
This proves the second claim of Lemma~\ref{xylem}.
\end{proof}



Let $t_0>0$, and $G_{t_0}$ be a $t_0$-relaxation of $G$. Hence,
\[
\begin{split}
\rbr{X_0,y_0}\in \arg\inf_{(X,y)\in \dd(G)}G_{t_0}(X,y)
&=\arg\inf_{(X,y)\in \dd(G)}t_0f_0(x)-\sum_{i=1}^{n+2}\psi_i(-f_i(X,y))\\
&=\arg\inf_{(X,y)\in \dd(G)}t_0\doubleOne^Ty-\ln \abs{X}-\ln \abs{I_d-X}- \sum_{i=1}^n \ln (y_i^2-\norm{Xp_i}_2^2)
\end{split}
\]
see Definition~\ref{relax}. 

\begin{proposition}
\label{complexity:ksm central path iterations bound}
Consider $\varepsilon\in\rbr{0,1}$, and let $x^*_{t_0}=\rbr{X_{t_0},{y_{t_0}}}$ satisfy $\lambda\rbr{G_{t_0},x^*_{t_0}}\leq0.01$. There exists an algorithm that computes an additive $\varepsilon$-approximation to the minimum of $\rlksm\rbr{k,\br{p_i}_{i=1}^n,w}$, referred to as Problem~\eqref{not:median_socp_rel}, in time
\begin{align}
    \label{complexity:sdp socp relaxation partial complexity}
    O\rbr{\rbr{n+d^2}^\omega\log_{\mu}\rbr{\frac{\sum_{i=1}^{n+2}\theta_i+1}{\eps t_0}}\rbr{\sum_{i=1}^{n+2}\theta_i\cdot\rbr{\mu-\log\mu}}}=O\rbr{\rbr{n+d^2}^\omega\rbr{n+d}{\ln{\frac{n+2d}{\varepsilon t_0}}}}.
\end{align}
\begin{proof}
 Recall that $\br{f_i}_{i=0}^{n+2}$ and $\br{\psi_i}_{i=1}^{n+2}$ represent sets of self-concordant functions. Given that $\br{f_i}_{i=1}^{n+2}$ consists of linear functions, it follows that $\br{\psi_i\rbr{-f_i}}_{i=1}^{n+2}$ forms a set of self-concordant functions. By Theorem~\ref{thm:cent}, taking $\mu=2$, $t\coloneq t_0$, and $x^*_{t}\coloneq x^*_{t_0}$, we conclude that there exists an algorithm that computes an additive $\varepsilon$-approximation to the minimum of Problem~\eqref{not:median_socp_rel} in \eqref{complexity:sdp socp relaxation partial complexity} time.
\end{proof}
\end{proposition}

We define an integer $\Delta$ so that $\ln{\Delta}$ is the word's size, i.e., the number of bits used to store entries in our finite-precision model. Thus, we assume that the absolute value of each non-zero entry of our input variables is between $\frac{1}{\Delta}$ and $\Delta$.

The following lemma provides a provably good initial point, for which, substituting the objective function of $G_{t_0}$, we obtain a bound on the distance from its minimum under its constraints. Thus, according to Lemma~\ref{newton}, the running time required to obtain $\rbr{X_{t_0},y_{t_0}}\in\dd\rbr{G}$ such that $\lambda\rbr{G_{t_0},\rbr{X_{t_0},y_{t_0}}}\leq0.01$ is bounded.

\begin{lemma}[Initial approximation\label{complexity:first iteration lemma}]
Let $P:=(p_i)_{i=1}^n$ be an ordered set of $n\geq1$ points in $\REAL^d$, and let $G(P)=(f_0,F,A,b)$ be its conic instance as defined in Lemma~\ref{xylem}. Let 
\begin{equation}
\label{complexity:t_0 definition}
    t_0:=\frac{2n}{\sum_{i=1}^n \sqrt{\norm{X_0p_i}_2^2+e}}.
\end{equation}
A pair $(X_{t_0},y_{t_0})\in \dd(G)$ that satisfies 
\begin{equation}\label{gtz}
\lambda\rbr{G_{t_0},\rbr{X_{t_0},y_{t_0}}}\leq0.01
\end{equation}
can be computed in $O\rbr{\rbr{n+d^2}^\omega\rbr{n+d\log{d}}}$ time.
%
%
%
\end{lemma}
\begin{proof}
Let $X_0:=\frac{d-k}{d}I_d$ be a matrix in $\REAL^{d\times d}$. 
Let $\tl_0:=(y_{0_1},\ldots,y_{0_n}) \in\REAL^n$ be a vector whose $i$th coordinate is
\begin{align}
    \label{complexity:initial y vector}
    {y_0}_i\coloneqq\sqrt{\norm{X_0p_i}_2^2+e},
\end{align}
for every  $i\in\cbr{n}$. Clearly $(X_0,y_0)\in \dd(G)$. 

Let 
\begin{equation}\label{optx}
(X^*,y^*):=(X_{t_0}^*,y^*_{t_0})\in \arg\inf_{(X,y)\in \dd(G)} G_{t_0}(X,y),
\end{equation}
and 
\[
\rho:=G_{t_0}(X_{0},y_{0})-G_{t_0}(X^*,y^*).
\]
Substituting $t:=t_0$ and $\tl_0:=(X_0,y_0)$ in Lemma~\ref{newton} yields that the desired pair $\tl_{t_0}:=(X_{t_0},y_{t_0})$ that satisfies
\[
\lambda\rbr{G_{t_0},\rbr{X_{t_0},y_{t_0}}}\leq 0.01,
\]
as in~\eqref{gtz}, can be computed in time 
\begin{equation}\label{nref}
O\rbr{\rho\cdot\rbr{n+d^2}^\omega}.
\end{equation}
It is left to bound
\begin{align}
\nonumber \rho&=G_{t_0}(X_{0},y_{0})-G_{t_0}(X^*,y^*)\\
\label{duf}&=    t_01^Ty_0-\sum_{i=1}^n\ln\rbr{{y_0^2}_i-\norm{X_0p_i}_2^2}-\ln\abs{X_0}
    -\ln\abs{I_d-X_0}\\
    &\quad-\rbr{t_01^Ty^*-\sum_{i=1}^n\ln\rbr{{y^*_i}^2-\norm{X^*p_i}_2^2}
    -\ln\abs{X^*}-\ln\abs{I_d-X^*}}\label{duf2},
\end{align}
where~\eqref{duf} and~\eqref{duf2} holds by substituting $(X,y):=(X_0,y_0)$ and $(X,y):=(X^*,y^*)$, respectively, in~\eqref{gtt}, together with $t:=t_0$. We now bound each expression as follows.

By \eqref{complexity:initial y vector},
\begin{equation}
    \label{complexity:initial socp bound}
    \sum_{i=1}^n\ln\rbr{{y_0^2}_i-\norm{X_0p_i}_2^2}=\sum_{i=1}^n\ln{e}=n.
\end{equation}
We have $\abs{X_0}=\rbr{\frac{d-k}{d}}^d$ and $\abs{I_d-X_0}=\rbr{\frac{k}{d}}^d$, and thus
\begin{align}
    \label{complexity:initial sdp bound}
    \ln\abs{X_0}+\ln\abs{I_d-X_0}=-d\ln{\frac{d^2}{k\rbr{d-k}}}\geq -d\ln (d^2)=-2d\ln d.
\end{align}
By the monotonicity of $\ln$, and since $\ln\rbr{x^2}\leq x$ for every $x>0$, respectively, we have
\begin{align}
    \label{complexity:optimum socp bound norm(X*pi)>=0 inequality}
    \sum_{i=1}^n\ln\rbr{{y^*_i}^2-\norm{X^*p_i}_2^2}&\leq\sum_{i=1}^n\ln\rbr{{y^*_i}^2}
    \leq 1^Ty^*
\end{align}

Since $X^*\in \dd(G)$, it satisfies $0 \preceq X^* \preceq I$, and thus its eigenvalues $\br{\sigma_i}_{i=1}^d$ are in $[0,1]$, and thus their multiplication is $\abs{X^*}\leq 1$. Similarly, the eigenvalues $\br{1-\sigma_i}_{i=1}^d$ of $I_d-X^*$ are in $[0,1]$. Hence,
\begin{align}
    \label{complexity:optimum sdp bound}
    \ln\abs{X^*}+\ln\abs{I_d-X^*}\leq\ln1+\ln1=0.
\end{align}

Plugging~\eqref{complexity:initial socp bound},~\eqref{complexity:initial sdp bound},~\eqref{complexity:initial sdp bound},~\eqref{complexity:optimum socp bound norm(X*pi)>=0 inequality} and~\eqref{complexity:optimum sdp bound} in~\eqref{duf2} and ~\eqref{duf2} yields
\begin{align}
    \rho&\leq t_01^Ty_0-n+\left(2d\ln d+1^Ty^*\right)\label{first iteration:bound sdp socp}.
\end{align}

Finally, we bound the last term $1^Ty^*$ by
\begin{align}\nonumber
    1^Ty^*\rbr{t_0-1}&=t_01^Ty^*-1^Ty^*\\
    &\leq    t_01^Ty^*-\sum_{i=1}^n\ln\rbr{{y^*_i}^2-\norm{X^*p_i}_2^2}\label{complexity:bound t_0w^Ty^*-w^Ty^*:1}\\
    &\leq    t_01^Ty^*-\sum_{i=1}^n\ln\rbr{{y^*_i}^2-\norm{X^*p_i}_2^2}-\ln{\abs{X^*}}-\ln{\abs{I_d-X^*}}\label{lnx}\\
    &\leq t_01^Ty_0-\sum_{i=1}^n\ln\rbr{{y_0^2}_i-\norm{X_0p_i}_2^2}
    -\ln{\abs{X_0}}-\ln{\abs{I_d-X_0}}\label{complexity:bound t_0w^Ty^*-w^Ty^*:2}\\
    &\leq t_01^Ty_0-n+2d\ln d ,\label{complexity:bound t_0w^Ty^*-w^Ty^*:3}
\end{align}
where~\eqref{complexity:bound t_0w^Ty^*-w^Ty^*:1} is established using \eqref{complexity:optimum socp bound norm(X*pi)>=0 inequality},~\eqref{lnx} holds by~\eqref{complexity:optimum sdp bound},~\eqref{complexity:bound t_0w^Ty^*-w^Ty^*:2} is by the optimality of $\rbr{X^*,y^*}$ in~\eqref{optx}, and the last inequality holds by~\eqref{complexity:initial socp bound} and~\eqref{complexity:initial sdp bound}.

Dividing~\eqref{complexity:bound t_0w^Ty^*-w^Ty^*:3} by $t_0-1$, and substituting in~\eqref{first iteration:bound sdp socp}  yields the final bound on~\eqref{duf},
\begin{equation}\label{rhoeq}
\begin{split}
    \rho&\leq t_01^Ty_0-n+\left(2d\ln d+1^Ty^*\right)\\
    &\leq t_01^Ty_0-n+\left(2d\ln d+\frac{t_01^Ty_0-n+2d\ln d}{t_0-1}\right)
    =n+ 2d\ln d+\frac{n+2d\ln d}{t_0-1}\leq2n+ 4d\ln d,
\end{split}
\end{equation}
where the last inequality holds since \[
t_0=\frac{2n}{1_n^Ty_0}\geq \frac{2n}{n\sqrt{e}}=\frac{2}{\sqrt{e}}>1.
\]
Substituting~\eqref{rhoeq} in~\eqref{nref} yields that the pair $(X_{t_0},y_{t_0})$ can be computed in 
\[
 O\rbr{\rbr{n+d^2}^\omega\rbr{n+d\ln{d}}}.
\]

\end{proof}

By Proposition~\ref{complexity:ksm central path iterations bound} and Lemma~\ref{complexity:first iteration lemma}, we can bound the total time complexity of our suggested $\KSM$ algorithm.
\begin{theorem}[running time\label{running}]
Let $k\in [d-1]$ and $\Delta\geq1$ be integers, and $P:=(p_i)_{i=1}^n\subseteq [-\Delta,\Delta]^d$. Let $S\subseteq\REAL^{d}$ be the output of a call to $\ksma\rbr{P,k}$; see Algorithm~\ref{alg:median_svd}. Then the projection matrix of $S$ can be computed in time
\begin{align}
\label{total complexity}
    O(n+d^2)^\omega(n+d)\ln \big(\Delta\cdot(n+2d)\big).
\end{align}
\end{theorem}
\begin{proof}
We notice that by \eqref{complexity:t_0 definition} and the definition of $\Delta$ we obtain that
\begin{equation}
    \label{t_0 bound}
    t_0\geq\frac{2n}{n\sqrt{d\cdot\Delta}}=\Omega\rbr{\frac{1}{\sqrt{d\cdot\Delta}}}.
\end{equation}
By substituting \eqref{t_0 bound} in Proposition~\ref{complexity:ksm central path iterations bound}, we obtain the given $\rbr{X_{t_0},y_{t_0}}$ calculated as in Lemma~\ref{complexity:first iteration lemma}, we obtain the the total running time for obtaining an additive $\varepsilon$ approximation to Problem~\eqref{not:median_socp_rel} is
\begin{equation*}
    O\rbr{\rbr{n+d^2}^\omega\rbr{\rbr{n+d}{\ln{\frac{\Delta\cdot\rbr{n+2d}}{\varepsilon d}}}+n+d\ln{d}}}=O\rbr{\rbr{n+d^2}^\omega\rbr{n+d}{\ln{\frac{\Delta\cdot\rbr{n+2d}}{\varepsilon}}}}.
\end{equation*}
\end{proof}

Our main result then follows by combining Theorem~\ref{running} and Theorem~\ref{alg:correctness theorem} as follows.
\begin{corollary}[$k$-subspace mean\label{maincol}]
\label{finaltheorem}
Let $k\in [d-1]$ and $\Delta\geq1$ be integers, and $P:=(p_i)_{i=1}^n$ be an ordered set of $n$ points in $[-\Delta,\Delta]^d$. Let $S\subseteq\REAL^{d\times d}$ be the output of a call to $\ksma\rbr{P,k}$; see Algorithm~\ref{alg:median_svd}. Then $S$ is a $\sqrt{d}$-approximation to the $k$-subspace median of $P$. 
Moreover, the projection matrix of $S$ can be computed in $(nd\ln \Delta)^{O(1)}$ time, i.e., polynomial in the size of the input $P$ and $k$.
\end{corollary}

\section{Experimental Results}
\label{experimental results}

\begin{figure*}[!hbt]
    \centering
        \setkeys{Gin}{width=0.45\linewidth}
    \subfloat[Log of sum of distances from "Statlog (Vehicle Silhouettes)"\label{vehicle dist}]{\includegraphics{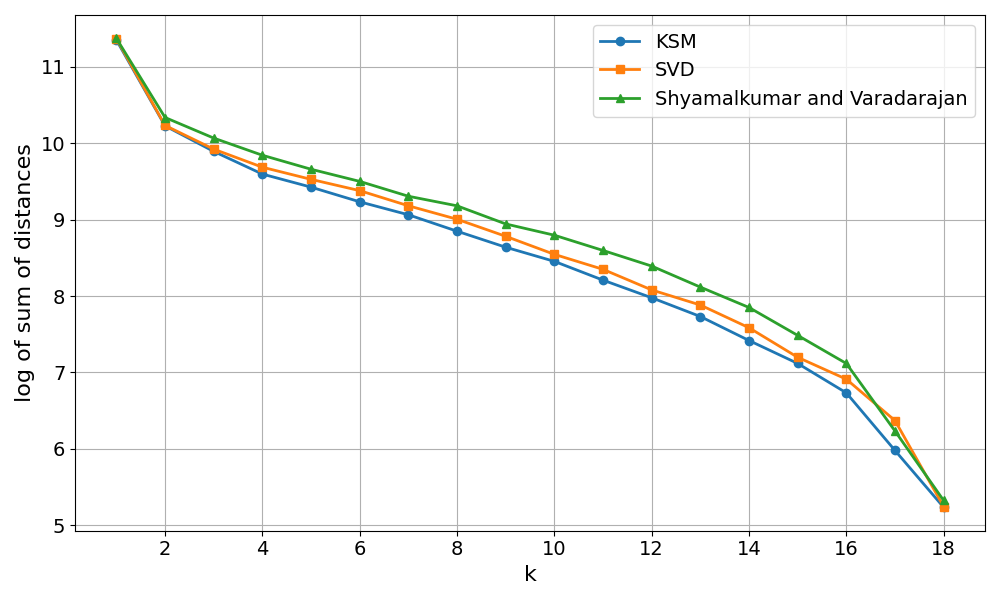}}
        \hfill
    \subfloat[Running time on "Statlog (Vehicle Silhouettes)"\label{vehicle time}]{\includegraphics{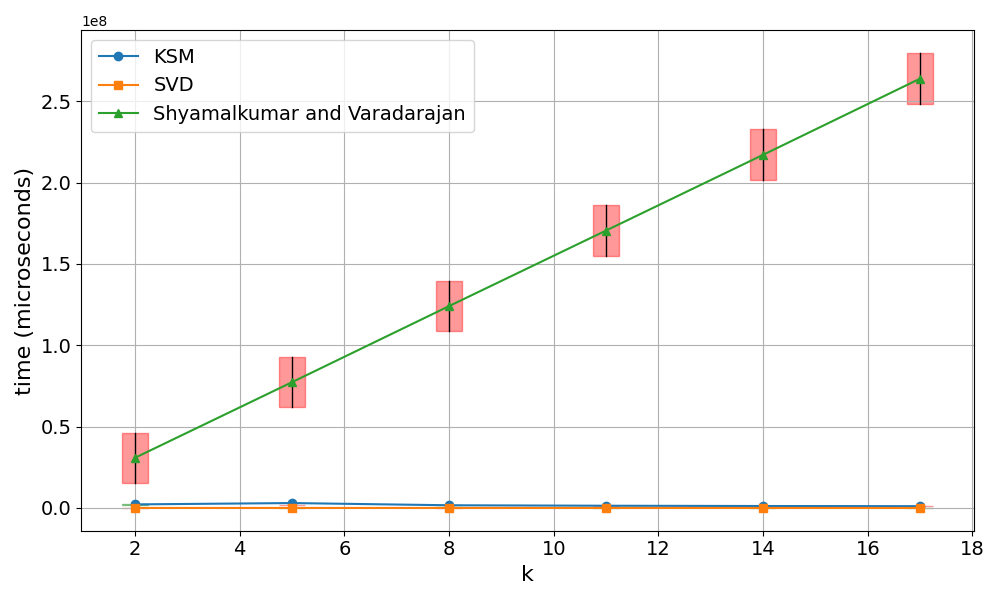}}
    \hfill
        \subfloat[Log of sum of distances from "Optical Recognition of Handwritten Digits"\label{optical dist}]{\includegraphics{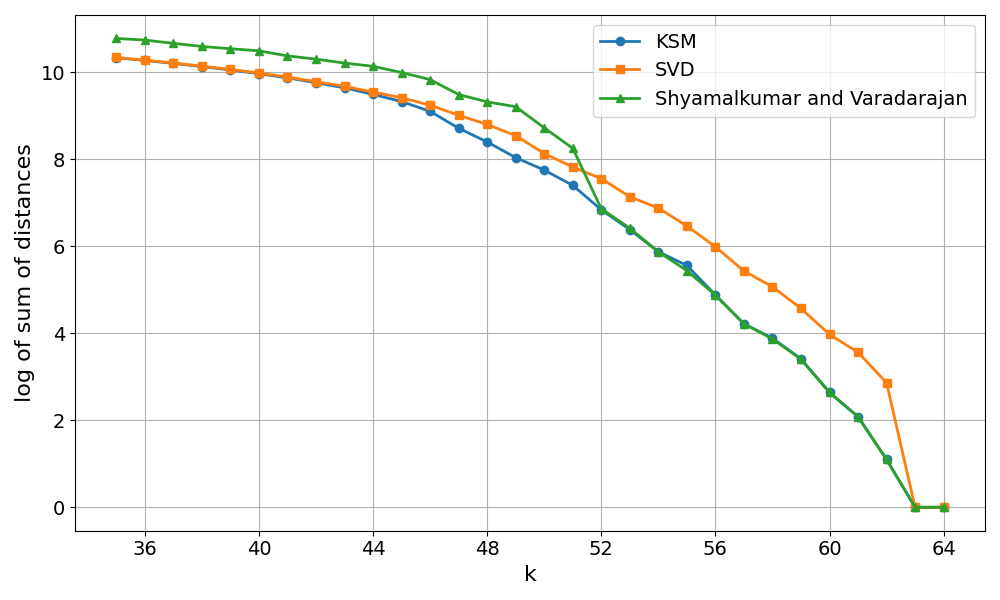}}
    \hfill
        \subfloat[Running time on "Optical Recognition of Handwritten Digits"\label{optical time}]{\includegraphics{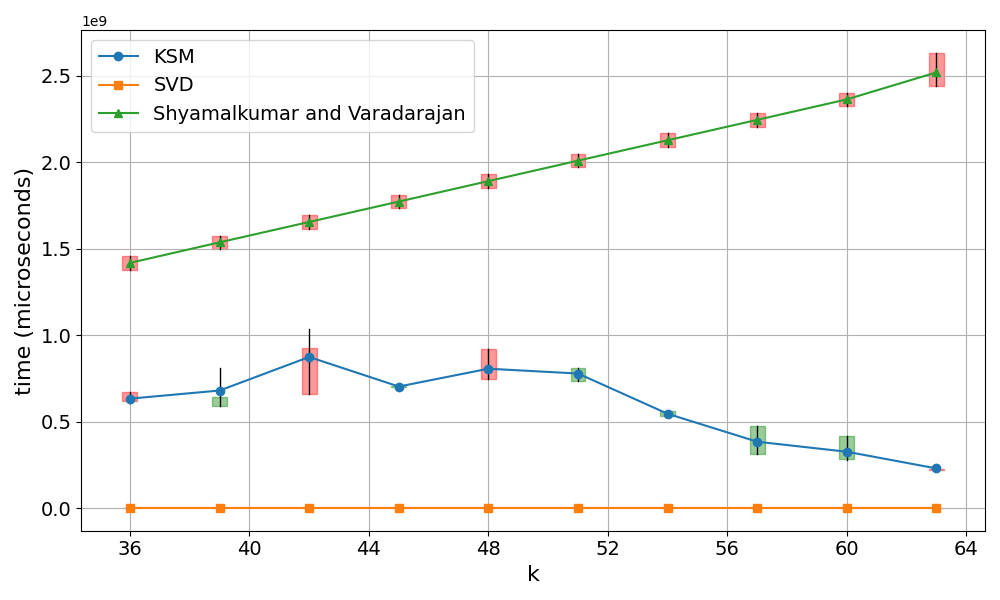}}
    \hfill
    \caption{Comparison of Algorithm~\ref{alg:median_svd} with SVD and the method by Shyamalkumar and Varadarajan on UC Irvine datasets. Figures~\ref{vehicle dist} and~\ref{optical dist}: Log of the sum of distances from the data points to the subspace. Figures~\ref{vehicle time} and~\ref{optical time}: Computation time in microseconds, shown with Japanese candlesticks and a batch size of 3.}
    \label{uci experiments}
\end{figure*}

We conducted experiments of dimensional reduction on two UC Irvine datasets: "Statlog (Vehicle Silhouettes)"~\cite{statlog_(vehicle_silhouettes)_149} ($n=847$ points in $d=19$ dimensions) and "Optical Recognition of Handwritten Digits"~\cite{optical_recognition_of_handwritten_digits_80} ($n=5621$ points in $d=65$ dimensions). As outlined by Siebert~\cite{Siebert1987VehicleRU} and Kaynak and Cenk~\cite{kaynak1995methods}, dimensional reduction constitutes an essential approach for constructing classifiers suitable for high-dimensional datasets. The experiments were carried out on an AWS EC2 instance of type r5b.4xlarge (16 vCPUs, 128GiB RAM).

Each experiment compares Algorithm~\ref{alg:median_svd} with the SVD Algorithm~\cite{golub1996matrix} and 100 iterations of the method suggested by Shyamalkumar and Varadarajan~\cite{shyamalkumar2007efficient}. We attempted to include the L1PCA algorithm~\cite{markopoulos2014optimal}, but available implementations~\cite{L1PCA_git} were infeasible for our datasets.

In Fig.~\ref{uci experiments}, the X-axis shows $k$, the dimension of the subspace. Figures~\ref{vehicle dist} and~\ref{optical dist} show the log of the sum of distances from the data points to the subspace, while figures~\ref{vehicle time} and~\ref{optical time}) show the computation time in microseconds, with Japanese candlesticks and a batch size of 3.

Our proposed algorithm demonstrates superior performance compared to both the SVD and the method introduced by Shyamalkumar and Varadarajan. As illustrated in Fig.~\ref{vehicle dist}, Algorithm~\ref{alg:median_svd} achieves distance sums that are reduced by a factor of up to 1.5 when compared to the SVD, and by a factor of up to 1.55 when compared to Shyamalkumar's Algorithm. Furthermore, in Fig.~\ref{optical dist}, Algorithm~\ref{alg:median_svd} presents sums that are diminished by up to 8.2 times relative to the SVD and by up to 3.2 times in relation to Shyamalkumar's Algorithm. Only for $k=55$, the method proposed by Shyamalkumar and Varadarajan achieves distance sums that are reduced by a factor of up to 1.13 compared to our algorithm. In summary, Algorithm~\ref{alg:median_svd} delivers mainly consistently superior results over both the SVD and a single iteration of the method proposed by Shyamalkumar and Varadarajan.

Figures~\ref{vehicle time} and \ref{optical time} present the computational run-time results. Although slower than SVD, the proposed method remains feasible. The method by Shyamalkumar and Varadarajan exhibits a high execution time relative to Algorithm~\ref{alg:median_svd}, even with only 100 evaluations. Attaining an $\rbr{1+\varepsilon}$-approximate solution necessitates exponential iterations, making it impractical for large datasets. The run-time of Algorithm~\ref{alg:median_svd} exhibits a slight decrease as $k$ increases, possibly due to its reliance on the starting point of the mixed SOCP-SDP problem~\eqref{not:median_socp_rel}. For further details, refer to Section~\ref{complexity analysis} and Theorem~\ref{running}.

\begin{figure*}[!hbt]
    \centering
    \includegraphics[width=0.8\linewidth]{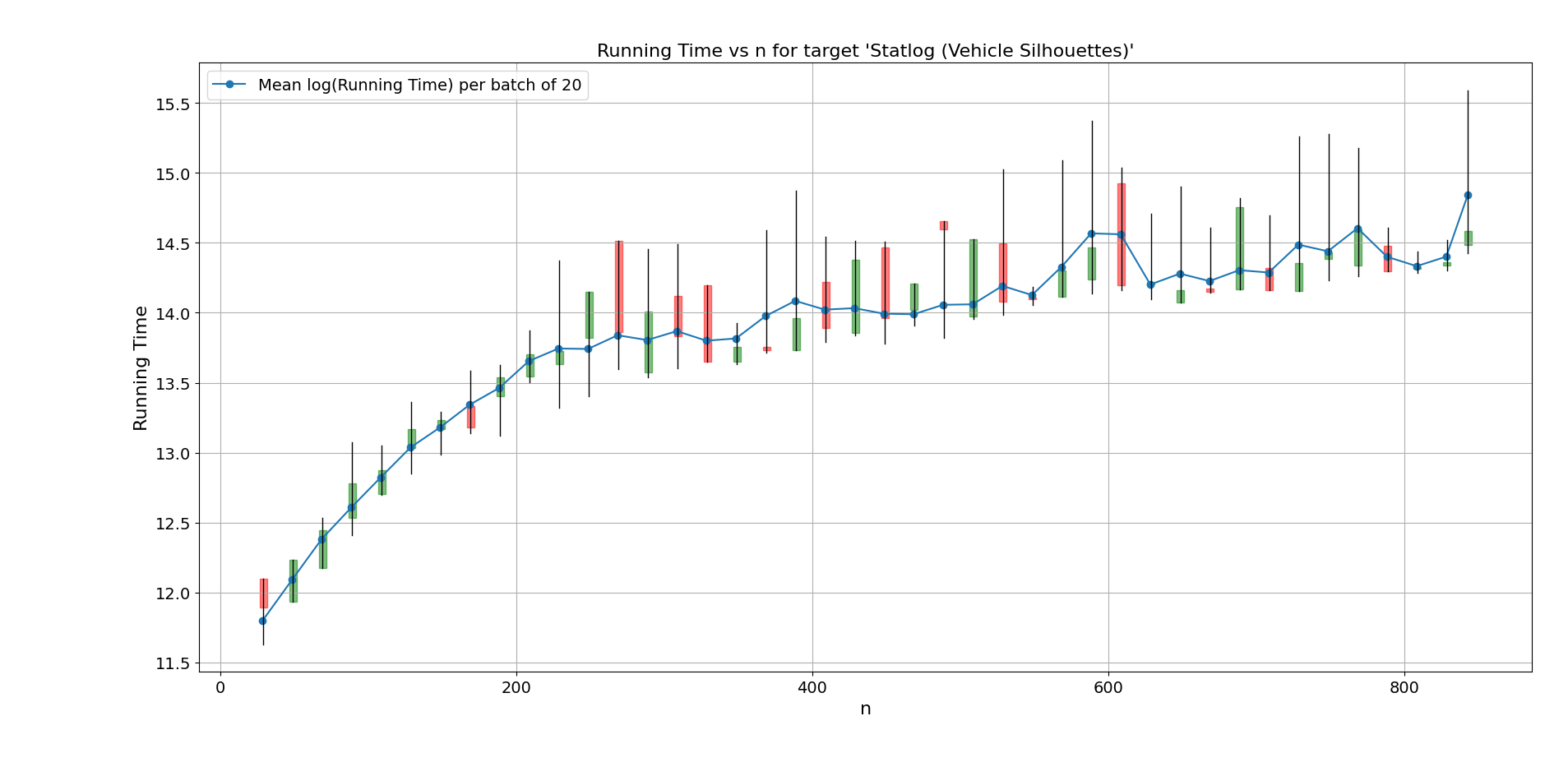}
    \caption{Logarithmic running time of Algorithm~\ref{alg:median_svd} on 'Statlog (Vehicle Silhouettes)' as the number of points increase (fixed $d=19$, $k=9$). The Japanese candlesticks are for 20-point batches.}
    \label{uci time analysis}
\end{figure*}

To validate the results we obtained in Section~\ref{complexity analysis}, we analyzed the runtime of the "Statlog (Vehicle Silhouettes)" dataset as the number of points varied, with $d=19$ and $k=9$ fixed. Fig.~\ref{uci time analysis} presents the results: the X axis shows the number of points, the Y axis the logarithmic running time of Algorithm~\ref{alg:median_svd} (in microseconds). The graph aligns with a logarithmic curve, affirming the polynomial complexity in $n$ and confirming Algorithm~\ref{alg:median_svd}'s polynomial runtime in $n$, as per Theorem~\ref{running}.

%

\appendix

\section{Related work\label{discussions}}
\paragraph{$k$-rank matrix approximation. }We now explain the relation between~\eqref{ab} and the $k$-subspace problem.
By letting $A\in\REAL^{n\times d}$ denote the matrix whose $i$th row is the $i$th point in $P$, in arbitrary order, the projections on the span of the columns of $X$ is $AX\in\REAL^{n\times k}$, and the projections in the original space are the rows of $AXX^T=A\Pi$ in $\REAL^{n\times d}$. Let $X_{\bot}\in\REAL^{n\times (d-k)}$ denote a matrix whose columns are orthonormal and orthogonal subspace to $X$, i.e., $V:=[X \mid X_{\bot}]\in\REAL^{d\times d}$ is an orthogonal matrix, and $XX^T+X_{\bot}X_{\bot}^T=\Pi+\Pi_{\bot}=V^TV=VV^T=I$. 

Defining $U:=AV=[AX \mid AX_{\bot}]\in\REAL^{n\times d}$ then yields a factorization 
\[
A=AVV^T=A(XX^T+X_{\bot}X_{\bot}^T)=A(\Pi+\Pi_{\bot})=A\Pi+A\Pi_{\bot}=UV^T,
\]
of $A$. Replacing the matrices $U$ and $V$ by their first $k$-columns $U_k$ and $V_k$ in $\REAL^{d\times k}$ yields a $k$-rank approximation $U_kV_k^T$ to $A$ in some sense that depends on the factorization. Given a $k$-subspace which is spanned by the columns of $V_k$, its closest point to an input point $p\in P$ is the projection of $p$ on the $k$-subspace, which is the corresponding row of $U_k$. Hence, $X=V_k$ that minimizes~\eqref{summ} above is the column space of a $k$-rank matrix $B$ that minimizes the ``mixed norm"
\begin{equation}\label{minb}
\min_{B} \norm{A-B}_{2,\zz}=\min_{U_k,V_k} \norm{A-U_kV_k}_{2,z}.
\end{equation}
Hence, the columns space of $B$ or $V_k$ is the same as the column space of the optimal solution $X$. This column space spans the $k$-subspace $S\subseteq\REAL^d$ that minimizes the sum of Euclidean distances to the power of $\zz$.

For the special case $z=2$, $V_k$ consists of the top $k$ right singular vectors of $A$, while the columns of $U_k$ are the left singular vectors after scaling. In this special case, $U_k$ has also mutually orthogonal columns, i.e. $U_k^TU_k=I$.
In other words, the optimal $k$-subspace implies an optimal $k$-rank approximation to a given matrix that depends on the given norm, which is a fundamental problem in fields such as linear algebra, machine learning and their applications.

\paragraph{Why $k$-subspace median?\label{why}} 
Although the $k$-subspace mean ($k$-PCA) can be computed efficiently, it lacks robustness when handling outliers or corrupted input, and the coordinates are usually not spared in any basis. A natural approach to handling these issues is to approximate the $k$-subspace median ($\KSM$), which aims to minimize the sum of (non-squared) Euclidean distances to the input points $(z=1$). To understand why the $\KSM$ demonstrates greater resilience to outliers and noise compared to $k$-PCA, it is useful to compare the mean with the median of the vector of distances to a subspace, which are numbers in $d=1$ dimensions~\cite{lopuhaa1991breakdown}. The mean is the value that reduces the total squared distances from the data points, whereas the median minimizes the total (non-squared) distances. If a single data point is moved toward infinity, the mean will also approach infinity. In contrast, to achieve a similar impact on the median, at least half of the data points must be shifted to infinity,  known as a \emph{breakdown point~\cite{lopuhaa1991breakdown}} of 50\%.

\paragraph{Affine subspace (flat)}. The $k$-subspace intersects the origin, by its definition. A straightforward extension for the $k$-subspace median (or any $\zz\geq 1$ norm) is the \emph{$k$-flat median} that minimizes the sum of distances among every \emph{affine} $k$-subspace of $\REAL^d$. That is, a translated $k$-subspace that does not necessarily intersect the origin. For the case $\zz=2$, it can easily be shown that the $k$-flat mean intersects the mean of the input set $P$. Hence, it is simply the mean $k$-subspace of $P$, after translating its points to their mean and then translating the resulting $k$-subspace back to the original coordinate system~\cite{pearson1901liii}. This is the main difference between $k$-PCA ($k$-flat mean) and $k$-SVD ($k$-subspace mean). However, for $\zz\neq2$, this reduction no longer holds. Instead, 
Maalouf et al. suggested a simple reduction from $k$-flat in $\REAL^d$ to $k$-subspace in $\REAL^{d+1}$ in~\cite{maalouf2020tight}. This is also why we consider only the $k$-subspace median, and not the $k$-subspace flat or affine subspace median in the rest of this section and paper.

\label{related work:provable algorithms}
\paragraph{Hardness.}
Clarkson and Woodruff~\cite[Theorem 2]{clarkson2015input} proved that it is NP-hard to approximate the $k$-subspace median of $n$ points in $\REAL^d$, up to a multiplicative factor of $\rbr{1+\sfrac{1}{d^{O\rbr{1}}}}$, if $k$ is part of the input, as in this paper. For the case $\ell_{1,1}$ exact solution that take time exponential in $n$ and then only $k$ were suggested in~\cite{markopoulos2013some,markopoulos2014optimal}, respectively.


\paragraph{Convex optimization.}
Our technique involves transforming the non-convex $k$-subspace median task into a convex optimization problem involving two cones. The first cone is the set of positive semi-definite (PSD) matrices in the vector space of symmetric matrices (see Fig.~\ref{fig:algorithm_illustration}), and the second cone is the second-order cone (SOC) in $\REAL^{d+1}$. The corresponding dual techniques are called SDP (semi-definite programming) and SOCP (second-order cone programming), and their combination is called SDP-SOCP~\cite{nesterov1994interior,mittelmann2003independent,molzahn2015mixed}.

\section{Related optimization functions\label{morerelated}}
The $k$-subspace median can be generalized to minimize the sum of Euclidean distances over every distance to the power of $\zz>0$~\cite{shyamalkumar2007efficient,ding2006r}. For $\zz\in \br{1,2,\infty}$, we obtain the $k$-subspace median (as in this paper), mean, and center, respectively.

\paragraph{The case $\zz=2$.} 
The optimal subspace for the common and easiest case $\zz=2$ is sometimes called the $k$-dimensional \emph{subspace mean} or $k$-SVD/PCA. It minimizes the sum of \emph{squared} Euclidean distances to the input points. The name comes from the fact that (a) the mean of $P$ (center of mass) minimizes the sum of the squared distances to the input points and (b) this subspace minimizes the mean or root mean squared (RMS) of the distances to the $n$ points. Similarly, the median minimizes the sum of non-squared distances, and the center minimizes the maximum distance (for $\zz=\infty$ norm). The names are also used in the context of $k$ centers, known as $k$-mean/median/center; see references in~\cite{feldman2007bi}.

It is often claimed that the optimal subspace for the case $\zz=2$ can be computed \emph{exactly} in $O(nd^2)$ time by taking the span of the largest $k$ singular vectors on the right of the $n\times d$ real matrix $A$ whose rows correspond to the input points~\cite{golub1996matrix,pearson1901liii}. These are the largest $k$ eigenvectors of $A^TA$, where ties are broken arbitrarily. This claim can be found in most related academic papers (see bibliography) and sources such as Wikipedia. However, computing the roots of a polynomial can be reduced to computing the eigenvalues of $A^TA$ via its companion matrix, which cannot be computed in closed-form (radicals) for $d>4$ by the Abel-Ruffini Theorem~\cite{ramond2022abel}. Hence, even for the simplest case $\zz=2$, we can only compute the approximation for the optimal subspace, without making any further assumptions about the input. 

Modern research~\cite{sarlos2006improved,har2014low,deshpande2006adaptive,deshpande2006matrix,drineas2006polynomial,drineas2006fast} has concentrated on developing algorithms that compute $k$-mean subspace in sub-quadratic time in $d$, achieving a $\rbr{1+\varepsilon}$-approximation to the optimal solution in expected time $O\rbr{nd\cdot\poly\rbr{k,\frac{1}{\varepsilon}}}$.

\paragraph{Precision.} After computing $A^TA$ in $O\rbr{nd^2}$ time, a multiplicative $(1+\eps)$-approximation to the $k$-subspace mean of $P$, for a given $\eps>0$, reduces to computing the eigenvalues of $A$ that takes time $O(d^3\log(1/\eps))$~\cite{pan1999complexity,stack}. For example, by substituting $\eps=\sfrac{1}{n^9}$ we can compute a $\rbr{1+\sfrac{1}{n^9}}$ approximation in $O(nd^2)$ time for a sufficiently large $n$. 
This approximation, which is only poly-logarithmic in $\sfrac{1}{\eps}$, is sometimes called \emph{machine precision}. Suppose that each coordinate at every input point $p\in P$ has only $b$ bits of precision for some integer $b\geq 1$. For example, $b=23$ for the float type in C or $b=52$ for double type. That is, after scaling (multiplying) the coordinates of $P$ by a sufficiently large integer, they are all integers in $\cbr{-2^b,2^b}$, that is, $P\subseteq \cbr{-2^b,\cdots,2^b}^d$.
In this case, the entries of the eigenvectors of $A^TA$ as its eigenvalues have also bounded precision $O\rbr{b}$, as proved e.g. in~\cite{feldman2020turning}. This implies that using $\eps\in \sfrac{1}{2^{\Omega(b)}}$ would give an exact solution for the $k$-subspace mean in $O\rbr{nd^2+d^2b^{O\rbr{1}}}$ time. In particular, the running time is $O\rbr{nd^2}$ time if each coordinate in $P$ has a finite precision $b\in O(1)$. This natural assumption in practice may justify the ignorance of this issue in so many texts. 

Another approach is to define a \emph{weakly polynomial time} algorithm~\cite{grotschel2012geometric} whose execution time is polynomially dependent on the input size in bits. In the context of the previous paragraph, if every coordinate of every input point can be represented via $b$ bits, then the size of the input in bits is $N\in O(ndb)$, and it can be stated that the running time $O\rbr{nd^2+d^2b^{O(1)}}=N^{O(1)}$ above is polynomial in the size of the input, without mentioning $b$. This situation is often common in convex optimization. Specifically, a strongly polynomial-time algorithm (which is independent of $b$) for linear programming has been cited as one of the 18 greatest unsolved problems of the 21st century~\cite{smale1998mathematical}.

\paragraph{The case $\zz=\infty$.}
This version of the problem is called \emph{The $k$-subspace center} due to the solution for the one-dimensional case $d=1$, as in the mean and median cases. In this problem, the subspace minimizes the maximum distance to the input points and is thus very sensitive to outliers, even more than the PCA. Varadarajan et al.~\cite{varadarajan2007approximating} showed a $O\rbr{\sqrt{\log n}}$-approximation polynomial time randomized algorithm. 

\paragraph{The case $\zz\in (2,\infty)$. }
Deshpande et al.~\cite{deshpande2011algorithms} use a convex relaxation of the rank constraint to the trace constraint. Unfortunately, their results do not apply to the $k$-subspace median, but only for the less robust cases where $p\geq2$. In addition, their algorithm is randomized.

\section{Data reduction}\label{datareduction}
There are many data reduction techniques such as coresets and sketches that do not suggest an approximation to variants of the $k$-subspace median problem, but instead suggest a reduction from the input set $P$ of $n$ points to a smaller number of weighted $m\ll n$ points (``coreset") or their linear combinations (``sketch"). However, there are very few algorithms for approximating the $k$-subspace median itself (possibly on the $m$ points). In fact, even the modern coresets that we are aware of focus on non-mixed $\ell_{\zz}:=\ell_{\zz,\zz}$ norms; see e.g. ~\cite{dasgupta2009sampling} and Section~\ref{datareduction}.

For a weighted set $\rbr{P,w}$ of input points, where $P\subseteq\REAL^d$, $\abs{P}=n$, and $w\in\REAL_{++}^n$, and a given approximation error $\varepsilon\in\rbr{0,1}$, a weighted set $\rbr{C,w_C}$ is a (strong) \emph{$\varepsilon$-coreset} for $(P,w)$ if $C\subseteq P$ and the sum of distances to any $k$-subspace to $P$, is the same, up to $1+\eps$ multiplicative error, to the sum of weighted distances to $C$.

Many have suggested efficient randomized algorithms to compute such a coreset to the $k$-subspace median; among them are Tukan et al.~\cite{tukan2020coresets}, Braverman et al.~\cite{braverman2016new}, and others~\cite{feng2021dimensionality,sohler2018strong,feldman2010coresets,feldman2020turning}. 
For $k=0$ (point-median), coresets were suggested in~\cite{cohen2021improved}. 

The construction of randomized coresets usually takes near-linear time in size $nd$ of the input and polynomial in $\frac{k}{\varepsilon}$. However, deterministic coresets are exponential in $k$, such as Sohler and Woodruff~\cite{sohler2018strong}. We emphasize that a coreset is a data-reduction technique that does not solve the $\KSM$ problem directly, as in this paper, but can be used to boost its performance to time that is linear or near-linear in the size of the input.

Sketches are similar to coresets, but instead of a weighted subset, each point in the sketch can be a linear combination of input points. Therefore, a sketch can be represented a as left multiplication of the $n\times d$ input matrix by a ``fat" sketch matrix. Sketches for operator Norm, Schatten norms, and subspace embeddings were suggested in~\cite{li2016tight}, with many references therein.

\section{Central Path Method\label{central}}
\label{complexity:central path method}
In this section, we describe the central path method to solve a convex optimization problem over $\REAL^n$ for $n\geq1$. The problem includes $d\geq1$ linear equality constraints and $m\geq1$ conic constraints, aiming for an additive $\varepsilon$-approximation of its optimum for a given $\varepsilon\in\rbr{0,1}$. In what follows, we can replace $\REAL^n$ with any space that is isomorphic to $\REAL^n$. In particular, this includes the Cartesian product of vector spaces or the space of $a\times b$ real matrices, where $a,b\geq1$ are integers.

\begin{definition}[proper cone~\cite{boyd2004convex}]
    \label{not:proper cone definition}
    The interior of a set $C\subseteq\REAL^n$ is denoted by
\begin{equation*}
    \interior{C}\coloneqq\br{x\in C\middle|\exists \varepsilon>0\colon \br{y\in\REAL^n\middle|\norm{x-y}_2\leq\varepsilon}\subseteq C}.
\end{equation*}
    A set $K\subseteq\REAL^n$ is called a \emph{proper cone} if it satisfies Properties (i)-(iii) as follows.
\begin{enumerate}
\renewcommand{\labelenumi}{(\roman{enumi})}
\item For every $x,y\in K$ and $\theta_1,\theta_2\in\REAL_{+}$ we have $\theta_1x+\theta_2y\in K$. 
\item If $x\in K$ and $-x\in K$ then $x=0$. 
    \item  $K$ is closed, and $\interior{K}\neq\emptyset$.
\end{enumerate}     
\end{definition}

\begin{definition}[logarithm barrier~\cite{boyd2004convex}]
    Let $k\geq1$ be an integer, $K\subseteq\REAL^{k}$ be a proper cone and $\theta>0$. A function $\psi:\interior{K}\rightarrow\REAL$ is a \emph{generalized logarithm} of degree $\theta\geq 0$ if it satisfies the following properties:
    \begin{enumerate}
\renewcommand{\labelenumi}{(\roman{enumi})}
        \item $\psi$ is concave, closed, twice continuously differentiable.
        \item Every $y\in \interior{K}$ and $s>0$ satisfy $\psi\rbr{sy}=\psi\rbr{y}+\theta\ln{s}$.
        \item Every $x\in\REAL^k$ satisfies $x^T\nabla^2\psi\rbr{y}x<0$, where $\nabla^2\psi\rbr{y}$ denotes the Hessian matrix of $\psi\rbr{y}$.
    \end{enumerate}
    Let $f:\REAL^n\to\REAL^{k}$ be a function that satisfies, for every $x,y\in\REAL^n$ and $\beta\in[0,1]$, 
\[
\beta-f(x)-(1-\beta)f(y)+f(\alpha x+(1-\beta)y)\in K.
\]
The $3$-tuple $(f,K,\psi)$ is a \emph{logarithmic barrier} in $\REAL^n$.
\end{definition}

\begin{definition}[$t$-relaxation\label{relax}]
Let $f_0:\REAL^n\to\REAL$, and $F=\br{f_i,K_i,\psi_i}_{i=1}^m$ be a set of $m\geq1$ logarithmic barriers in $\REAL^n$. The \emph{degree} of $F$ is $\deg(F):=\sum_{i=1}^m \theta_i$,  where $\theta_i$ is the degree of $f_i$, for $i\in[m]$. 
For a given matrix $A\in\REAL^{d\times n}$, and a vector $b\in\REAL^n$ we define $G:=(f_0,F,A,b)$ as the \emph{conic instance} for the optimization problem $\inf_{x\in \dd(G)} f_0(x)$ over the domain
\begin{equation}
\begin{split}
    \min_{x\in\REAL^n} \quad &  f_0(x)    \\
    \text{s.t.} \quad & Ax=b
    \\
    & -f_i(x)\in K_i \quad  \forall i\in\cbr{n}.
\end{split}
\end{equation}

For $t\geq 0$, we define the \emph{$t$-relaxation} of $G$ to be the function $G_{t}:\dd(G)\to\REAL$ that maps every $x\in \dd(G)$ in the domain
\[
\dd(G):=\bigcap_{i\in [m]} \br{x\in \REAL^n\mid -f_i(x)\in \interior{K_i} \text{ and }Ax=b}
\]
to 
\[
G_{t}(x):=tf_0(x)-\sum_{i=1}^m \psi_i\big(-f_i(x)\big).
\]
\end{definition}

\begin{definition}[self-concordant\label{selfproblam}]
    Let $H\subseteq\REAL$. A 3-continuously differentiable function $f:H\rightarrow\REAL$ is \emph{self-concordant} if for every $x\in H$, we have
    \begin{equation*}
        \abs{f'''\rbr{x}}\leq2{f''\rbr{x}}^{\frac{2}{3}}.
    \end{equation*}
    More generally, for an integer $n\geq1$ and $H\subseteq\REAL^n$, a function $h:H\rightarrow\REAL$ is self-concordant if for every $x\in H$ and $v\in\REAL^n$, the function $g_{x,v}$,
    that maps every $t\in\br{t\in\REAL\middle|x+vt\in H}$ to $g_{x,v}\rbr{t}=h\rbr{x+vt}$, is self-concordant.
The conic instance $G:=(f_0,F,A,b)$ is \emph{self-concordance}, if $$\psi(-f):\br{x\in\REAL^n\mid -f(x)\in \interior {K}}\to\REAL$$ is self-concordance, for every logarithmic barrier $(f,K,\psi)\in F$, and $f_0$ is also self-concordance.    
\end{definition}

In what follows, $\psi_i(-f_i):\br{x\in\REAL^n\mid -f(x)\in\interior{K_i}}$ is the composition $\psi_i\circ (-f_i)$ of the functions $\psi_i$ and $-f_i$ that maps to $\psi_i(-f_i(x))$ every $x\in\REAL^n$ that satisfies $-f(x)\in \interior{K_i}$.

\begin{definition}[Newton's Decrement~\cite{nesterov1994interior}]
    Let $G=\rbr{f_0,F,A,b}$ be a self-concordance conic instance and $t>0$. The \emph{Newton's Decrement} of $G_t$ at $x\in\dd\rbr{G}$ is
    \begin{equation*}
        \lambda\rbr{G_t,x}=\sqrt{\rbr{\nabla G_t\rbr{x}}^T\rbr{\nabla^2G_t\rbr{x}}^{-1}\rbr{\nabla G_t\rbr{x}}}.
    \end{equation*}
\end{definition}

\begin{proposition}[~\cite{nesterov1994interior}]
    Let $G=\rbr{f_0,F,A,b}$ be a self-concordance conic instance,  $t>0$ and $y\in\dd\rbr{G}$. If $\lambda\rbr{G_t, y}\leq0.01$ then
    \[
    \inf_{x\in\dd\rbr{G}}G_t\rbr{x}\geq G_t\rbr{y}+\lambda^2\rbr{G_t,y}.
    \]
\end{proposition}

\begin{lemma}[Reduction to $t$-relaxation~\cite{nesterov1994interior}]
\label{t-relaxation lemma}
Let $G=(f_0,F,A,b)$ be a self-concordance conic instance, $t>0$ and $x_t\in\dd\rbr{G}$. 
If $\lambda\rbr{G_t,x_t}\leq0.01$ then
\[
f_{0}(x_t)
\leq \inf_{x\in \dd(G)} f_0(x)+ O\left(\frac{\deg(F)+1}{t}\right).
\]
\end{lemma}

\begin{lemma}[Inner Newton iterations~\cite{nesterov1994interior}\label{newton}]
Let $G:=(f_0,F,A,b)$ be a self-concordance conic instance in $\REAL^n$, $t>0$,  and $\tl_0\in\dd\rbr{G}$. There is an algorithm that gets $(G,t,\tl_0)$ as input, and outputs $\tl_t\in \dd(G)$ that satisfies $\lambda\rbr{G_t,\tl_t}\leq0.01$.
For every $\mu>1$, its running time is
\[
n^{\omega}\deg\rbr{F}\cdot O\rbr{\mu-1-\log\mu},
\]
if $\lambda\rbr{G_{\frac{t}{\mu}},\tl_0}\leq 0.01$, and 
\[
 n^{\omega}\cdot O\rbr{G_{t}(\tl_0)-\inf_{x\in\dd\rbr{G}}G_{t}(x)}.
\]
otherwise. 


\end{lemma}
\newcommand{\sss}{s}
\begin{algorithm}[!hbt]
\DontPrintSemicolon
\caption{\pathalg$\rbr{G,\mu,t_0,x_0,\eps}$}
\label{central path algorithm}
    \SetKwInOut{Input}{Input}
    \SetKwInOut{Output}{Output}
    \Input{A self concordance conic instance $G=(f_0,F,A,b)$ in $\REAL^n$, $\mu>1$, an integer $t_0>0$, and $\eps\in(0,1)$.}       
    \Output{$x_{\iter}\in \dd(G)$.}
    \vspace{3mm}
Let $$\sss\in \Theta\left(\log\rbr{\frac{\deg\rbr{F}+1}{\eps t_0}}/\log \mu\right)$$ be an integer whose exact value can be determined from the proof of Lemma~\ref{t-relaxation lemma}.\\
\For {$j:=1$ to $\sss$}{
Compute $x_{j}\in \dd(G)$ that satisfies $\lambda\rbr{G_{\mu^jt_0},x_j}\leq0.01$\\
\tcp{Possibly by substituting $z_0\coloneq x_{j-1}$ in Lemma~\ref{newton}.}
}
\Return $x_\sss$   
\end{algorithm}


Following Lemma~\ref{t-relaxation lemma} and Lemma~\ref{newton}, $\pathalg\rbr{G,\mu,t_0,\eps}$ (Algorithm~\ref{central path algorithm}) calculates an additive $\eps$-approximation to $\inf_{x\in\dd\rbr{G}}f_0\rbr{x}$ for self concordance conic instances.

\begin{theorem}[Central path method~\cite{nesterov1994interior}\label{thm:cent}]
Let $G=(f_0,F,A,b)$ be a self-concordance conic instance, $\mu>1$, $t_0>0$, and $\eps\in (0,1)$. 
Let $x_0\in \dd(G)$ such that $\lambda\rbr{G_{t_0},x_0}\leq0.01$, and $x^*$ be the output of a call to $\pathalg\rbr{G,\mu,t_0,x_0,\eps}$; see Algorithm~\ref{central path algorithm}.
    Then $x^*\in \dd(G)$ and
\[
f_0(x^*)
\leq \eps+\inf_{x\in \dd(G)} f_0(x).
\]
Moreover, $x^*$ can be computed in time
\[
n^\omega\cdot \deg\rbr{F}\cdot O\rbr{\log\rbr{\frac{\deg\rbr{F}+1}{\eps t_0}}}.
\]
\end{theorem}

%

\bibliography{main}
\bibliographystyle{ieeetr}

\end{document}